\documentclass[letterpaper]{article}
\usepackage{uai2020}
\usepackage[margin=1in]{geometry}

\usepackage{times}

\usepackage[round]{natbib}

\usepackage{amsmath}
\usepackage{amssymb}
\usepackage{graphicx}
\usepackage{enumitem}
\usepackage{booktabs}
\usepackage{wrapfig}
\usepackage{comment}

\usepackage{appendix}
\usepackage{url}

\usepackage{pgfplots, pgfplotstable}
\pgfplotsset{compat=newest}
\pgfplotsset{colormap/blackwhite}
\usepgfplotslibrary{groupplots}
\usepgfplotslibrary{dateplot}
\usetikzlibrary{math}
\usetikzlibrary{positioning}
\usetikzlibrary{patterns}

\definecolor{col1}{HTML}{8D230F}

\usepackage{xspace}
\makeatletter
\DeclareRobustCommand\onedot{\futurelet\@let@token\@onedot}
\def\@onedot{\ifx\@let@token.\else.\null\fi\xspace}
\def\eg{\emph{e.g}\onedot} 
\def\ie{\emph{i.e}\onedot}

\makeatother

\newcommand{\fref}[1]{Fig.~\ref{#1}}

\newcommand{\R}{\mathbb{R}}
\newcommand{\KL}[2]{\text{KL}[#1 \, || \, #2]}

\newcommand{\E}{\mathbb{E}}
\newcommand{\GP}[0]{\mathcal{GP}}
\newcommand{\Data}{\mathcal{D}}

\newcommand{\y}{\mathbf{y}}
\newcommand{\x}{\mathbf{x}}

\newcommand{\m}{\mathbf{m}}
\newcommand{\mmu}{\boldsymbol{\mu}}
\newcommand{\mSigma}{\boldsymbol{\Sigma}}
\newcommand{\tmmu}{\tilde{\boldsymbol{\mu}}}
\newcommand{\bmmu}{\bar{\boldsymbol{\mu}}}
\newcommand{\bmsigma}{\bar{\boldsymbol{\sigma}}^2}
\newcommand{\tmSigma}{\tilde{\boldsymbol{\Sigma}}}
\newcommand{\U}{\mathbf{u}}
\newcommand{\F}{\mathbf{f}}
\newcommand{\Fb}{\bar{\mathbf{f}}}
\newcommand{\Fz}{\mathbf{f}^\mathbf{z}}
\newcommand{\Fs}{\mathbf{f}^*}
\newcommand{\X}{\mathbf{x}}
\newcommand{\Xs}{\mathbf{x}^*}
\newcommand{\Y}{\mathbf{y}}
\newcommand{\Sb}{\mathbf{S}}
\newcommand{\Z}{\mathbf{z}}
\newcommand{\N}{\mathcal{N}}
\newcommand{\given}{\: | \:}
\newcommand{\dd}{\mathrm{d}}

\newcommand{\xn}{\Fb_{\ell - 1}}
\newcommand{\fn}{\F_\ell}
\newcommand{\s}{\boldsymbol{\sigma}_\ell^2}
\newcommand{\eps}{\varepsilon_{\ell - 1}}
\newcommand{\D}[1]{(#1)'}
\newcommand{\DD}[1]{(#1)''}
\newcommand{\V}[1]{\mathbb{V}\text{ar}\left[#1\right]}
\newcommand{\EE}[1]{\mathbb{E}\left[#1\right]}
\newcommand{\sn}{\boldsymbol{\sigma}_{\text{noise}}^2}
\newcommand{\q}{\mathbf{q}}
\newcommand{\K}{\mathbf{K}}

\newcommand{\RR}{\mathbb{R}} 

\DeclareMathOperator{\Tr}{Tr}

\newcommand\blfootnote[1]{%
  \begingroup
  \renewcommand\thefootnote{}\footnote{#1}%
  \addtocounter{footnote}{-1}%
  \endgroup
}

\title{Compositional uncertainty in deep Gaussian processes}

\author{
  Ivan Ustyuzhaninov*\\
  University of T\"ubingen \\
  \And
  Ieva Kazlauskaite* \\
  University of Bath,\\
  Electronic Arts \\
  \And
  Markus Kaiser \\
  Siemens AG,\\
  TU Munich \\ 
  \AND
  Erik Bodin \\
  University of Bristol \\
  \And
  Neill D. F. Campbell \\
  University of Bath, \\
  Royal Society \\
  \And
  Carl Henrik Ek \\
  University of Bristol \\
} 

\begin{document}

\maketitle

\begin{abstract}
Gaussian processes (GPs) are nonparametric priors over functions. Fitting a GP implies computing a posterior distribution of functions consistent with the observed data. Similarly, deep Gaussian processes (DGPs) should allow us to compute a posterior distribution of compositions of multiple functions giving rise to the observations. However, exact Bayesian inference is intractable for DGPs, motivating the use of various approximations. We show that the application of simplifying mean-field assumptions across the hierarchy leads to the layers of a DGP collapsing to near-deterministic transformations. We argue that such an inference scheme is suboptimal, not taking advantage of the potential of the model to discover the compositional structure in the data. To address this issue, we examine alternative variational inference schemes allowing for dependencies across different layers and discuss their advantages and limitations.\blfootnote{* Equal contribution}
\end{abstract}

\section{INTRODUCTION}
\label{sec:intro}
\begin{figure*}[h]
    \begin{center}
        \includegraphics[width=0.85\linewidth]{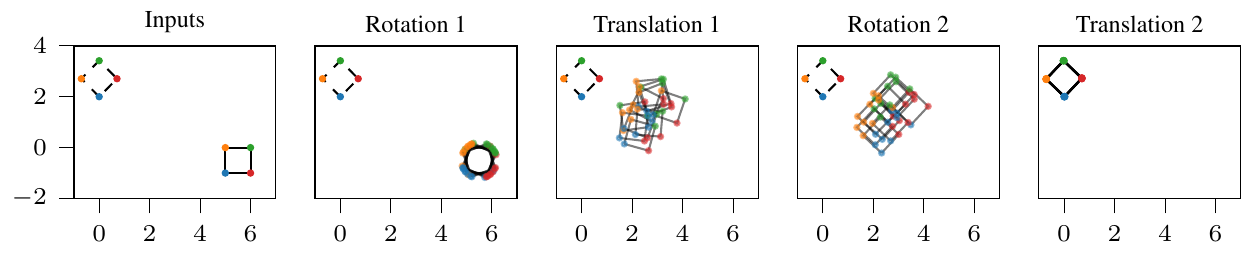}
    \end{center}
    \vspace{-0.5cm}
    \caption{Compositional model (toy example): the transformation of the solid rectangle onto the dashed one is decomposed as $T_2 \circ R_2 \circ T_1 \circ R_1$ where $R_i$ and $T_i$ are rotations and translations. Different sampled realisations of these transformations are overlaid, showing the \emph{compositional uncertainty}. Approximating $R_i$ and $T_i$ as independent transformations does not allow us to capture such uncertainty, collapsing to a single realisation of the composition.}
    \label{fig:rotated_squares_example}
\end{figure*}

Hierarchical learning studies functions represented as compositions of other functions, $f = f_L \circ \ldots \circ f_1$.
Such models provide a natural way to model data generated by a hierarchical process, as each $f_\ell$ represents a certain part of the hierarchy, and the prior assumptions on $\{f_\ell\}_{\ell=1}^{L}$ reflect the corresponding prior assumptions about the data generating process.
DGPs~\citep{Damianou:2013}, which are compositions of GPs, allow us to impose explicit prior assumptions on $\{f_\ell\}$ by choosing the corresponding kernels. 
Since different compositions can fit the data equally well (see an illustration in \fref{fig:rotated_squares_example}), DGPs are inherently unidentifiable, and this lack of identifiability should be captured by an adequate Bayesian posterior, allowing us to quantify uncertainties pertaining to each $f_\ell$. We refer to this uncertainty as \emph{compositional uncertainty}. This uncertainty can be thought of as the epistemic uncertainty~\citep{DerKiureghian:2009, Gal:2016} 
describing how the layers of the hierarchy jointly compose the observed data.

While the DGP posterior captures compositional uncertainty, exact Bayesian inference in DGPs is intractable~\citep{Damianou:2013}.
In this work we show that the typically used approximate inference schemes impose strong simplifying assumptions, making intermediate DGP layers collapse to deterministic transformations.\footnote{In cases where the data has high observational noise, the noise is explained by introducing an uncertainty in one or multiple layers of the composition. We focus on the case where the data is noiseless, thus the uncertainty in each of the layers arises only due to the ambiguity in the compositional structure.} 
This corresponds to representing a DGP as a single-layer GP with a transformed kernel~\citep{Dunlop:2018}, similar to GPs with kernels parametrised by a deterministic function (\eg a neural network). Such behaviour might not be a problem in practice if the goal is to design a model that only provides a high marginal likelihood of the data, however, it does not make full use of the capacity of DGP as it fails to describe the uncertainty that stems from the potential decomposition in the hierarchy.
Distributions over compositions, and the resulting compositional uncertainty, are important for applications, \eg for temporal alignment of time series data~\citep{Kaiser:2018, Kazlauskaite:2019}, in reinforcement learning~\citep{Jin:2017} as well as for building more interpretable models where each layer in the hierarchy expresses a meaningful functional prior~\citep{Sun:2019}.

We address the issue of collapsing compositional uncertainty by proposing variational distributions and corresponding inference methods that explicitly model the dependencies between the layers, resulting in variational posteriors that capture compositional uncertainty. 
So doing, we highlight the limitations of existing approaches and lay the ground for future work in uncertainty quantification in DGPs.
Our main contributions are: 
\begin{itemize}[leftmargin=1.5em,itemsep=0pt,topsep=-1pt]
    \item We demonstrate that variational distributions over the inducing points that are factorised \emph{across layers} lead to a collapse of compositional uncertainty,
    \item We provide an intuitive as well as a quantitative argument for this behaviour by drawing a link between the work on mean-field variational inference for DGPs and the models of regression with noisy inputs~\citep{Girard:2003}; 
    \item We propose modifications to the factorised variational distribution that incorporate the dependencies between the inducing points in different layers, and discuss the corresponding inference procedures,
    \item We use the proposed variational inference approaches to further illustrate how the correlations across the layers are necessary in order to argue about compositional uncertainty.
\end{itemize}

The remainder of the paper is structured as follows.
We first provide a background to DGPs with an emphasis on approximate inference and discuss the method of \citep{Salimbeni:2017} in detail, using it as the starting point for our argument on the collapse of compositional uncertainty, presented in Sec.~\ref{sec:mean_field}.
In Sec.~\ref{sec:beyond_mean_field} we propose variational distributions that aims to address the shortcomings of the layer-wise factorisation.
In Sec.~\ref{sec:experiments} we illustrate the behaviour of the proposed methods and discuss potential areas of applications.
\section{BACKGROUND: MODELS OF DGPs}
\label{sec:background}

\paragraph{Previous work} The hierarchical GP construction was originally motivated from the perspective of latent variable models~\citep{Lawrence:2004} and was designed with a specific application in mind. In the early work on DGPs, \citet{Lawrence:2007} proposed a model that captured the hierarchical structure in the human skeleton, that allowed to produce interpretable generative models of human motion. However, most of the later work shifted the emphasis from uncovering specific interpretable hierarchical structures to employing a hierarchical construction to design models that are more flexible than a standard GP (in particular, by weakening the assumptions about a joint Gaussian structure in the observations). For example, \citet{Lazaro:2012} proposed a hierarchical (two-layer) GP model to allow for non-stationary observations. \citet{Damianou:2013} drew further parallels between DGPs and deep belief networks, and proposed a DGP construction beyond two layers for both supervised and unsupervised settings. Concurrently, the MAP estimation used in the early works~\citep{Lawrence:2007} was replaced with variational inference schemes, initially proposed for the latent variable model~\citep{Titsias:2010} and later adapted for the hierarchical DGP setting~\citep{Damianou:2013}. 

However, the variational inference approach of~\citet{Damianou:2013} was shown to be prohibitive for large data sets, motivating further research on inference schemes that scale to large data sets~\citep{Hensman:2013, Hensman:2014, Dai:2016, Bui:2016, Gal:2016:2, Hensman:2017, Salimbeni:2017, Cutajar:2019}. 
A different line of thought emerges from the work on inference using stochastic gradient Hamiltonian Monte Carlo~\citep{Havasi:2018}. 
The authors recognise the issue of compositional uncertainty, highlighting the fact that most of the existing (variational) approaches to inference are limited to estimating single modes of the posterior distributions in each layer of the hierarchy. 
As inference using MC is typically very costly, the authors note that it is beneficial to decouple the model in terms of the inducing points for the mean and the variance, which results in a highly non-convex optimization problem that requires careful parameterisation to improve the stability of convergence. Various issues with numerical stability, poor convergence and underestimation of uncertainty have also been reported in the context of variational approximations \citep{Hensman:2014, Kaiser:2018}. \citet{Duvenaud:2014} show a pathological behaviour of the concentration of density along a single dimension as the number of layers increases, and propose including direct links between the inputs and each individual layer.

\paragraph{Doubly stochastic variational inference (DSVI)} Our work builds on the variational approximation scheme introduced by \citet{Salimbeni:2017}, thus here we provide a short recap of the main ideas from this work and introduce the notation that is used throughout the rest of this paper. 
Given a dataset\footnote{Throughout the paper we consider one-dimensional data but the general considerations also apply in many dimensions.} $\Data = \{(x_j, y_j)\}_{j=1}^J$, with $x_j, y_j \in \R$, we model $y_j = (f_L \circ \ldots \circ f_1)(x_j)$, where $f_\ell \sim \GP(\mu_\ell(\cdot), k_\ell(\cdot, \cdot))$. 
We denote the inputs as $\X = (x_1, \ldots, x_J) \in \R^J$, and the evaluations of the intermediate layers at the entire vector of inputs $\X$ as $\F_\ell \sim (f_\ell \circ \ldots \circ f_1)(\X)$ for $\ell = 2, \ldots, L$. 
The DGP joint distribution is
\begin{equation}
    p(\Y, \F_L, \ldots, \F_1 \given \X) =
        p(\Y \given \F_L) \prod_{\ell=1}^L p(\F_\ell \given \F_{\ell-1}),
    \label{eq:dgp_joint}
\end{equation}
where $p(\F_\ell \given \F_{\ell-1}) \sim \GP(\mu_j(\F_{\ell-1}), k_j(\F_{\ell-1}, \F_{\ell-1}))$ is a GP prior for the $\ell$-th layer, and we define $\F_0 = \X$. Integrating $\{\F_\ell\}$ from \eqref{eq:dgp_joint} to obtain a marginal likelihood is intractable, since that requires integrating a product of Gaussian factors, each of which contains $\F_\ell$ inside a non-linear kernel.

To overcome this limitation, variational inference is used to estimate the lower bound on \eqref{eq:dgp_joint}. To this end, each DGP layer $\ell$ is augmented with $M$ inducing locations $\Z_\ell \in \R^M$ and inducing points $\U_\ell \in \R^M$, resulting in the following augmented joint distribution:
\begin{equation}
    \begin{aligned}
        p(\Y, & \{\F_\ell\}, \{\U_\ell\} \given \X, \{\Z_\ell\}) =
            p(\Y \given \F_L) \times \\
            & \times \prod_{\ell=1}^L
                p(\F_\ell \given \F_{\ell-1}, \U_\ell, \Z_{\ell-1}) 
                p(\U_\ell \given \Z_{\ell-1}),
        \label{eq:dgp_augmented_joint}
    \end{aligned}
\end{equation}
where $p(\F_\ell \given \F_{\ell-1}, \U_\ell, \Z_{\ell-1}) \sim \N(\mmu_\ell, \mSigma_\ell)$ is a GP posterior at inputs $\F_{\ell-1}$ given values of $\U_\ell$ at $\Z_{\ell-1}$. 
The specific form of $\mmu_\ell$ and $\mSigma_\ell$ is as follows (note a slight abuse of notation: $\mu_\ell(\cdot)$ is a mean function, while $\mmu_\ell$ is a posterior mean):
\begin{align*}
    \mmu_\ell & = \mu_\ell(\F_{\ell-1}) + \alpha_\ell(\F_{\ell-1})^T (\U_\ell - \mu_\ell(\F_{\ell-1})), \\
    \mSigma_\ell & = k_\ell(\F_{\ell-1}, \F_{\ell-1}) - \alpha_\ell(\F_{\ell-1})^T k_\ell(\Z_{\ell-1}, \Z_{\ell-1}) \, \alpha_\ell(\F_{\ell-1}), \nonumber
\end{align*}
where 
\begin{equation}
\alpha_\ell(\F_{\ell-1}) = k_\ell(\Z_{\ell-1}, \Z_{\ell-1})^{-1} k_\ell(\Z_{\ell-1}, \F_{\ell-1}).
\label{eq:dsvi_alpha_definition}
\end{equation}

Introducing a factorised variational distribution over the inducing points
\begin{equation}
    q(\{\U_\ell\}) = q(\U_1) \ldots q(\U_L), \,\,\,\,\, q(\U_\ell) \sim \N(\m_\ell, \Sb_\ell)
    \label{eq:dsvi_mean_field_variational_distribution}
\end{equation}
the likelihood lower bound is as follows:
\begin{equation}
    \begin{aligned}
    \mathcal{L}(\Y) & \geq \E_{q(\F_L)} [ \log p(\Y \given \F_L)] -  \\
    & \quad - \sum_{\ell = 1}^L \KL{q(\U_\ell)}{p(\U_\ell \given \Z_{\ell-1})}.
    \end{aligned}
    \label{eq:dsvi_bound_expectation}
\end{equation}
A key insight of \citet{Salimbeni:2017} is that the expectation in \eqref{eq:dsvi_bound_expectation} can be efficiently estimated by a Monte-Carlo estimator. This is possible by marginalising the inducing points $\{\U_\ell\}$ from the variational posterior, obtaining
\begin{equation}
    \begin{aligned}
        q(\F_L, \ldots, \F_1)
        & = \prod_{\ell=1}^L \int p(\F_\ell \given \U_\ell) q(\U_\ell) \, \dd\U_\ell \\
        & = q(\F_L \given \F_{L-1}) \ldots q(\F_1 \given \X),
    \end{aligned}
    \label{eq:integral_U_DSVI}
\end{equation}
with $q(\F_\ell \given \F_{\ell-1}) \sim \N(\tmmu_\ell, \tmSigma_\ell)$, where
\begin{align}
    \tmmu_\ell & = \mu_\ell(\F_{\ell-1}) + \alpha_\ell(\F_{\ell-1})^T (\m_\ell - \mu_\ell(\F_{\ell-1})), \label{eq:dsvi_mu_tilde} \\
    \tmSigma_\ell & = k_\ell(\F_{\ell-1}, \F_{\ell-1}) - \label{eq:dsvi_sigma_tilde} \\
        & \quad - \alpha_\ell(\F_{\ell-1})^T (k_\ell(\Z_{\ell-1}, \Z_{\ell-1}) - \Sb_\ell) \, \alpha_\ell(\F_{\ell-1}). \nonumber
\end{align}

The bound in \eqref{eq:dsvi_bound_expectation} can be estimated by sequentially sampling from  $q(\F_\ell \given \F_{\ell-1})$ using \eqref{eq:dsvi_mu_tilde} and \eqref{eq:dsvi_sigma_tilde}. The time complexity of this step is linear in the number of data points, since each marginal $[\F_\ell]_j$ can be drawn independently (we only need marginals of the final layer $\F_L$ in \eqref{eq:dsvi_bound_expectation}).
\section{MEAN-FIELD DGPs}
\label{sec:mean_field}

\begin{figure}[t]
    \centering
    \includegraphics{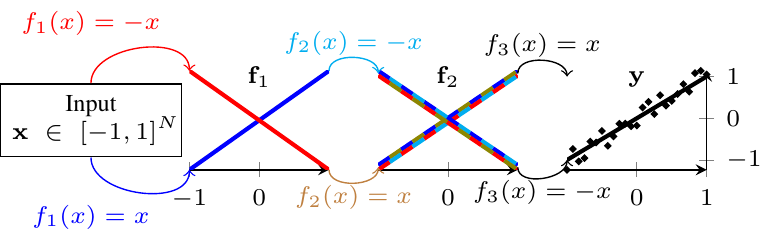}
    \vspace{-0.5cm}
    \caption{A toy example illustrating three layer compositions where each layer is either $f_\ell(x) = x$ or $f_\ell(x) = -x$. Multiple compositions map $\X$ to $\Y$, this uncertainty is illustrated by showing the range of different values of $\F_1 = f_1(\X)$ and $\F_2 = f_2(\F_1)$. If a variational distribution over $\{f_\ell\}$ is factorised, the posterior compositions collapse to a single realisation.}
    \label{fig:toy_example}
\end{figure}

In this section we argue that factorised variational distributions of inducing points, \eg \eqref{eq:dsvi_mean_field_variational_distribution}, imply that the layers in a DGP collapse to deterministic transformations. 

\subsection{INTUITION}

If a DGP $f_L \circ \ldots \circ f_1$ maps fixed inputs $\X$ to fixed outputs $\Y$, the functions $\{f_\ell\}$ must be dependent, because every realisation of this composition must map the \emph{same} $\X$ to the \emph{same} $\Y$. This is illustrated in \fref{fig:toy_example}, which shows a composition of three layers, each of which could either be $f_\ell(x) = x$ or $f_\ell(x) = -x$. Depending on the choices of $f_1$ and $f_2$, the input is mapped by $f_2 \circ f_1$ to one of the two realisations of $\F_2$ (as shown by the colour code in the corresponding panel), and $f_3$ must be chosen in such a way that $\F_2$ is mapped to $\Y$. 
Therefore, in this example, $f_3$ depends on the choice of $f_1$ and $f_2$. 
However, if $\{f_\ell\}$ were independent, then the only way to ensure that every realisation of the composition fits the data would be for each layer to implement a deterministic transformation (\ie either $f_\ell(x) = x$ or $f_\ell(x) = -x$ such that there are zero or two instances of $f_\ell(x) = -x$). 
Another illustration of this idea is provided in \fref{fig:rotated_squares_example}, in which movement of a square is represented as a composition of correlated rotations and translations, allowing us to see a variety of possible movements. 
However, a model with independent transformations would converge to a single possible sequence of rotations and translations.

The same intuition holds for general DGPs. Analogously to choosing either $x$ or $-x$ in \fref{fig:toy_example}, inducing locations $\Z_\ell$ and points $\U_\ell$ define the transformation implemented by the corresponding layer through the predictive posterior $p(\F_\ell \given \F_{\ell-1}, \U_\ell, \Z_{\ell-1})$. Following a similar argument, the DGP layers collapse to deterministic transformations to ensure good data fits unless they are dependent to allow multiple different compositions to fit the data.

\subsection{QUANTITATIVE ARGUMENT}

Assume that the DGP layers $\{f_\ell\}$ are independent.
Then the distribution of the outputs of layer $\ell - 1$ can be thought of as uncertain inputs\footnote{Regression models that include input uncertainty can generally be formulated as: $\y = f(\x + \varepsilon_\x)$, where $\y$ are observations, $\x$ are noise-free inputs and $\varepsilon_\x$ is zero-mean noise.} to the layer $\ell$.
Similarly to DGPs, the inference in such models is complicated by the need to propagate a distribution through a non-linear mapping.
Such models have been studied in the context of GP regression~\citep{Girard:2003, Mchutchon:2011, Bijl:2018} and have also been discussed in relation to DGPs~\citep{Damianou:2015}, though not in the context of compositional uncertainty.  

Assuming for simplicity that our dataset consists of a single point, \ie $\Data = \{(x, y)\}$, we can write $\F_{\ell} = (f_{\ell} \circ \ldots \circ f_1)(x) = f_\ell(\F_{\ell - 1}) = f_\ell(\Fb_{\ell - 1} + \eps)$, with $\Fb_{\ell - 1}$ as the mean\footnote{We use bold notation for $\Fb_{\ell - 1}$, even though it refers to a scalar, to distinguish it from the notation we use for functions.} of $\F_{\ell - 1}$ and $\eps$ as an appropriate zero-mean noise (not necessarily Gaussian, since marginals of DGP layers are not Gaussian in general~\citep{Damianou:2015}), the variance of which we denote as $\sn := \V{\eps}$. 
We want to show that the variance of $\F_\ell$ increases with increasing variance of $\eps$, which would imply that unless the layers collapse, \ie $\eps = 0$, there is finite variance in the final layer $\F_L$. That constitutes a poor fit to observations that contain low observational noise (noiseless in the limit), forcing the layers to collapse to deterministic transformations.

High observational noise might lead to the layers not collapsing despite being independent. However, such uncertainty is the observational noise spread across the layers, rather than compositional uncertainty due to multiple compositions explaining the data. To make our arguments clearer, we assume noiseless observations.

\paragraph{Linear approximation} We can approximate $\fn$ as
\begin{equation}
    \fn = f_\ell(\F_{\ell - 1}) \approx f_\ell(\xn) + \eps \, f'_\ell(\xn),
\end{equation}
where $f_\ell(\xn) \sim p(\F_\ell \given \xn, \U_\ell, \Z_{\ell-1}) = \N(\bmmu_\ell, \bmsigma_\ell)$, with $\bmmu_\ell$ and $\bmsigma_\ell$ given in \eqref{eq:dsvi_mu_tilde} and \eqref{eq:dsvi_sigma_tilde}. Note that both $\bmmu_\ell$ and $\bmsigma_\ell$ are functions of $\xn$, which we omit to not clutter the notation; the derivatives below are taken w.r.t. $\xn$.

\begin{figure*}[t]
    \centering
    \includegraphics{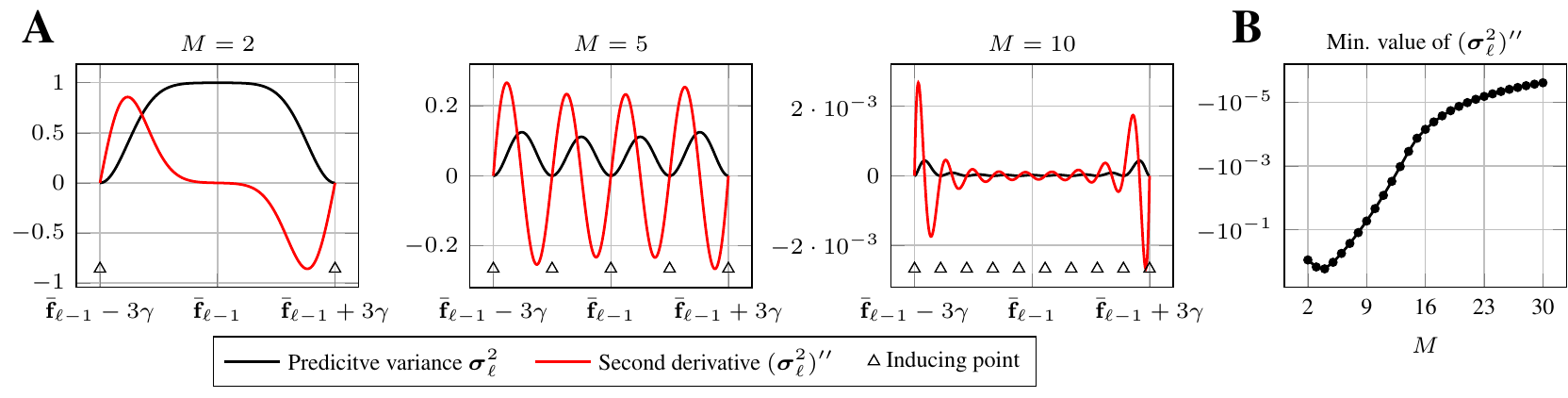}
    \vspace{-0.7cm}
    \caption{\textbf{A:} Predictive posterior variance $\s$ and its second derivative $\DD{\s}$ of $\ell$-th layer in a $3\gamma_\ell$-neighbourhood $\Delta_\ell$ of the noiseless input $\xn$ for different numbers of inducing points. \textbf{B:} Minimum value of $\DD{\s}$ as a function of number of inducing points $M$.}
    \label{fig:second_derivative}
\end{figure*}

The evaluation of a GP and its derivative are jointly distributed as follows \citep{Rasmussen:2005}:
\begin{equation}
    \begin{bmatrix}
        f_\ell(\xn) \\
        f'_\ell(\xn)
    \end{bmatrix}
    \sim
    \N \left(
        \begin{bmatrix}
            \bmmu_\ell \\
            \bmmu'_\ell
        \end{bmatrix},
        \begin{bmatrix}
            \s & \D{\s} \\
            \D{\s} & \DD{\s}
        \end{bmatrix}
    \right).
    \label{eq:joint-gp-and-derivative}
\end{equation}

Similarly to~\citet{Mchutchon:2011}, we compute a linear transformation of \eqref{eq:joint-gp-and-derivative}
and obtain that
\begin{align*}
    \EE{\fn \given \eps} & = \bmmu_\ell + \eps \bmmu'_\ell, \\
    \V{\fn \given \eps} & = \s - 2 \eps \D{\s} + \eps^2 \DD{\s}.
\end{align*}

Using the law of total variance we have
\begin{equation*}
\begin{aligned}
    \V{\fn} &= \EE{\V{\fn | \eps}} + \V{\EE{\fn | \eps}}, \\
& \hspace{-2.2cm} \text{where} \\
    \EE{\V{\fn | \eps}} 
        &= \s + \sn \, \DD{\s}, \\
    \V{\EE{\fn | \eps}} 
            & = \V{\bmmu_\ell + \eps \, \bmmu'_\ell} = \sn \cdot \left( \bmmu'_\ell \right)^2.
\end{aligned}
\end{equation*}

Combining these results together we obtain
\begin{equation}
    \hspace{-0.13cm}
    \V{\fn} = \s + \sn \left[ \left( \bmmu'_\ell \right)^2 + \DD{\s} \right] + \, O(\eps^2).
    \label{eq:noisy-variance}
\end{equation}

The only term in \eqref{eq:noisy-variance} that can be negative is $\DD{\s}$, potentially making the variance of the GP output at a noisy input smaller that the variance at a fixed input (i.e. $\s$).

\paragraph{Counterexample} Such an example can indeed be constructed. \citet{Girard:2003} study GPs with uncertain inputs and compute an exact expression for $\V{\F_\ell}$ as a function of $\sn$ assuming $\eps$ is Gaussian. 
Assuming there is a single inducing point $\U_\ell = 0$, the derivative of $\V{\F_\ell}$ is negative at 0 provided that $\Fb_{\ell - 1}$ is sufficiently far away from $\U_\ell$ (in comparison to the length scale; see the derivation in the Appendix~\ref{app:counterexample}). 
This means that the input noise might \emph{reduce} the output variance. 
However, such an example relies on the inputs to the $\ell$-th layer, $\Fb_{\ell - 1}$, appearing in the regions of the input space that are poorly covered by the inducing points. 
Consequently, this scenario only occurs if the inducing points are placed in a way that leads to a poor fit of the observed data. 

\paragraph{Inducing points limit} The counterexample above relies on a degenerate setting in which the inducing points are far from the observations. Here we consider a limiting case that corresponds to a more realistic situation of sufficiently many inducing points near $\F_\ell$ (the limit of arbitrary many inducing points is a conceptually desirable setting, complicated by the computational constraints). Specifically, in each layer we assume $M$ linearly spaced inducing points $\Z_\ell = \{z_{\ell-1}^1, \ldots, z_{\ell-1}^M\}$ in $\Delta_\ell := [\xn - 3 \gamma_\ell, \xn + 3 \gamma_\ell]$,
where $\gamma_\ell$ is the kernel length scale in layer $\ell$. This assumption means that the input to each layer is contained in an interval $\Delta_\ell$ covered by the inducing points.

The behaviour of \eqref{eq:noisy-variance} under such an assumption is illustrated in \fref{fig:second_derivative}. The minimum value of $\DD{\s}$ approaches zero as $M$ increases; this suggests that the input noise leads to increased predictive posterior variance apart from degenerate cases of inducing points not covering the input region corresponding to the observed inputs $\x$, and the predictive mean derivative $\left( \bmmu'_\ell \right)^2$ being sufficiently small (\ie the function implemented by the $\ell$-th layer being close to a constant one).

To summarise, we argue that under the assumption of $\F_{\ell} = (f_{\ell} \circ \ldots \circ f_1)(x)$ being contained in an interval covered by the inducing locations $\Z_\ell$ for the next layer, the variance in $\F_{\ell}$ leads to increased variance in $\F_{\ell + 1}$, and hence in $\F_L$. 
Therefore, for $\F_L$ to fit a noiseless observation $y$, the variance in intermediate layers has to be reduced, implying that the layers collapse to deterministic transformations.
\section{BEYOND FACTORISED VARIATIONAL DISTRIBUTIONS}
\label{sec:beyond_mean_field}

To further investigate the effect of the factorisation imposed by the mean-field variational inference, we propose two alternative variational inference schemes that allow for correlations between layers. 
By relaxing the mean-field assumption across layers, we aim to uncover a range of solutions that are consistent with the data and follow the prior belief about each of the individual layers.
In Sec.~\ref{subsec:joint_gaussian_inducing_points} we present a natural generalisation of a factorised variational distribution \eqref{eq:dsvi_mean_field_variational_distribution} to capture the marginal dependencies between the layers. In Sec.~\ref{subsec:variational_intermediate_layers} we present an alternative variational approximation that introduces dependencies between the layers by linking the inducing points and locations of the neighbouring layers.

\subsection{JOINTLY GAUSSIAN INDUCING POINTS}
\label{subsec:joint_gaussian_inducing_points}

A straightforward modification of the DSVI variational approximation (Sec.~\ref{sec:background}) allowing us to capture the dependencies between the layers is to introduce correlations between the inducing points by modelling them with a jointly Gaussian variational distribution:
\begin{equation}
    q(\U_1, \ldots, \U_L) \sim \N(\m, \Sb),
    \label{eq:variational_distribution_joint_Gaussian}
\end{equation}
with $\m \in \RR^{LM}, \, \Sb \in \RR^{LM \times LM}$. The variational posterior is then given by
\vspace{-0.2cm}
\begin{equation}
    \hspace{-2pt} 
    q(\{\F_\ell\}, \{\U_\ell\}) = q(\U_1, \ldots, \U_L) \prod_{\ell = 1}^L p(\F_\ell \given \F_{\ell-1}, \U_\ell).
    \label{eq:variational_posterior_joint_Gaussian}
\end{equation}
The corresponding likelihood lower bound has the same structure as \eqref{eq:dsvi_bound_expectation} with the KL term, $\KL{q(\U_1, \ldots, \U_L)}{p(\U_1) \ldots p(\U_L)}$, that can be computed in closed form (it involves two Gaussians). The expectation $\E_{q(\F_L)}[\log p(\Y \given \F_L)]$ is, however, harder to estimate in case of variational distribution \eqref{eq:variational_distribution_joint_Gaussian}. The integral \eqref{eq:integral_U_DSVI} no longer factorises into a product of integrals, which means that we can no longer integrate $\{\U_\ell\}$ out from $q(\{\F_\ell\}, \{\U_\ell\})$ and draw samples from $q(\F_L)$ in the same way as in \citep{Salimbeni:2017}. We consider two approaches to address this issue.

\paragraph{Sampling $\{\U_\ell\}$} 
We start by noting that, conditioned on $\{\U_\ell\}$, we can draw samples from $q(\F_L)$ in the same way as in \citep{Salimbeni:2017}. Specifically, to estimate $\E_{q(\F_L)}[\log p(\Y \given \F_L)]$, we
\begin{enumerate}[leftmargin=1.5em,itemsep=0pt,topsep=-1pt]
    \item Draw $S$ samples \\
    $\{(\U_1^s, \ldots, \U_L^s)\}_{s=1}^S \stackrel{iid}{\sim} q(\U_1, \ldots, \U_L)$,
    \item For each sample $(\U_1^s, \ldots, \U_L^s)$, draw $\protect{\F_L^s \sim q(\F_L \given \U_1^s, \ldots, \U_L^s)}$ by recursively drawing from $p(\F_\ell \given \F_{\ell-1}, \U_\ell^s)$, which are regular GP posterior distributions conditioned on $\U_\ell^s$,
    \item Compute a Monte Carlo estimate \\ $\E_{q(\F_L)}[\log p(\Y \given \F_L)] \approx \frac{1}{S} \sum\limits_{s} \log p(\Y \given \F_L^s)$.
\end{enumerate}
This approach is easy to implement and it can be applied in a variety of settings (\eg when  $q(\{\U_i\})$ is not Gaussian, as long as we can sample from it and reparametrise the gradients). 
However, that comes at the cost of introducing another sampling step, resulting in $\E_{q(\F_L)}[\log p(\Y \given \F_L)]$ being estimated by two nested Monte-Carlo estimators, implying an increased overall variance of the estimator and the need to carefully choose the appropriate number of samples~\citep{Rainforth:2019}.
Moreover, drawing coherent samples from $q(\U_1, \ldots, \U_L)$ has computational complexity of $O(L^3 M^3)$ leading to an overall complexity of $O(L^3 M^3 + LNM^2)$ per estimation.

\paragraph{Analytic marginalisation} 
To address statistical and computational limitations of the above method, we propose another approach consisting of analytically integrating $\{\U_\ell\}$ from \eqref{eq:variational_posterior_joint_Gaussian}.
To do so we assume that $q(\{\U_\ell\})$ admits a chain-like factorisation, namely
\begin{equation}
    q(\{\U_\ell\}) = q(\U_L \given \U_{L-1}) \ldots q(\U_2 \given \U_1) q(\U_1).
    \label{eq:jointly_Gaussian_chain_constraint}
\end{equation}

\begin{wrapfigure}[7]{r}{0.42\columnwidth}
    \vspace{-10pt}
    \includegraphics{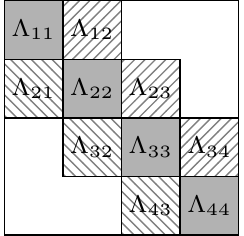}
    \vspace{-10pt}
    \caption{Precision matrix $\Lambda$ induced by \eqref{eq:jointly_Gaussian_chain_constraint}.}
    \label{fig:precision_matrix_block_tridiagonal}
\end{wrapfigure}
The precision matrix across all layers, $\protect{\Lambda = \Sb^{-1} \in \RR^{LM \times LM}}$, encodes the conditional independence assumptions, and \eqref{eq:jointly_Gaussian_chain_constraint} implies that such matrix is block-tridiagonal (Fig.~\ref{fig:precision_matrix_block_tridiagonal}). 
The advantage of this assumption is that the number of parameters in the unconstrained $\Sb$ scales quadratically with the number of layers, while \eqref{eq:jointly_Gaussian_chain_constraint} implies a linear growth.
Assuming that the variational distribution \eqref{eq:variational_distribution_joint_Gaussian} satisfies the factorisation \eqref{eq:jointly_Gaussian_chain_constraint}, we analytically marginalise $\{\U_\ell\}$ from the variational posterior \eqref{eq:variational_posterior_joint_Gaussian}, obtaining
\begin{equation}
    \int q(\{\F_\ell\}, \{\U_\ell\}) \,\dd\{\U_\ell\} = \prod_{\ell=1}^L p(\F_\ell \given \F_{\ell-1}, \ldots \F_1, \X),
    \label{eq:jointly_Gaussian_integrated_U}
\end{equation}
where $p(\F_\ell \given \F_{\ell-1}, \ldots \F_\ell, \X) \sim \N(\tmmu_\ell, \tmSigma_\ell)$ with the mean and the covariance are defined recursively as follows:
\begin{equation}
\begin{aligned}
    \tmmu_1 & = \mu_1(\X) + \alpha_1(\X)^T (\m_1 - \mu_1(\X)) \\
    \tmSigma_1 & = k_1(\X, \X) - \alpha_1(\X)^T (k_1(\Z_1, \Z_1) - \Sb_{11}) \, \alpha_1(\X) \nonumber
\end{aligned}
\end{equation} and $\alpha_1(\X)$ is defined in \eqref{eq:dsvi_alpha_definition}. 
For $i > 1$, $\tilde{\mu}_i$ and $\tilde{\Sigma}_i$ are recursively defined as
\begin{align}
    \tmmu_\ell = \,\, & \mu_\ell(\F_\ell) + \alpha_\ell(\F_{\ell-1})^{T} (\m_\ell + \Sb_{\ell, \ell-1} \, \alpha_{\ell-1}(\F_{\ell-1}) \times \nonumber  \\ 
    & \times \tmSigma_{\ell-1}^{-1} (\F_{\ell-1} - \tmmu_{\ell-1} - \alpha_{\ell-1}(\X)^T \times \label{eq:integrated-mean} \\
    & \times (\m_{\ell-1} - \mu_{\ell-1}(\Z_{\ell-1})) - \mu_\ell(\Z_\ell)), \nonumber
\end{align}
\begin{equation}
\begin{aligned}
    \tmSigma_\ell = & \,\, k_\ell(\F_{\ell-1}, \F_{\ell-1}) - \alpha_\ell(\F_{\ell-1})^T (k_\ell(\Z_\ell, \Z_\ell) -  \\
    & - \Sb_{\ell\ell} + \Sb_{\ell,\ell-1} \, \alpha_{\ell-1}(\F_{\ell-1}) \tmSigma_{\ell-1}^{-1} \times \\
    & \times \alpha_{\ell-1}(\F_{\ell-1})^T \Sb_{\ell-1, \ell}) \alpha_\ell(\F_{\ell-1}),
\end{aligned}\label{eq:integrated-var}
\end{equation} where $\Sb_{ij} = \text{cov}(\U_i, \U_j)$.

The derivation is provided in the Appendix~\ref{app:analytic-marginalisation}. Using these results, $\E_{q(\F_L)}[\log p(\Y \given \F_L)]$ can be estimated analogously to DSVI by recursively sampling $\F_i$ using \eqref{eq:jointly_Gaussian_integrated_U}.

\subsection{INDUCING POINTS AS INDUCING LOCATIONS}
\label{subsec:variational_intermediate_layers}
In this section we discuss an alternative variational approximation, that connects the inducing points and the inducing locations of the neighbouring layers. Instead of directly modelling the inducing points in every layer, we only consider the inducing inputs $\Z$ in the first layer and variational distributions over $\{\Fz_\ell \sim (f_\ell \circ \ldots \circ f_1)(\Z)\}$. The advantage of such an approach is that unlike the variational distributions of inducing points, the factorisation of a variational distribution over $\{\Fz_i\}$ does not imply that the variational posterior collapses to a single realisation of a composition fitting the data. 
In such a setting, $\Fz_{\ell-1}$ and $\Fz_\ell$ can be thought of as inducing pairs of the $\ell$-th layer, meaning that the inducing points of a previous layer are the inducing locations of the next one.

\paragraph{Intuition} Let us revisit the illustration given in Fig.~\ref{fig:toy_example}. Assuming for this example that $\Z = \X$, we independently sample values of $\F_1$ and $\F_2$ from $q(\F_1)q(\F_2)$ (\ie one of the two types of coloured lines in panels $\F_1$ and $\F_2$). Given such a sample, we can deduce the functions $f_1, f_2, f_3$. For example, the colour of $\F_1$ denotes the choice of $f_1$, the second colour of $\F_2$ (the first colour is that of $f_1$) corresponds to $f_2$, and $f_3$ is chosen to map $\F_2$ to the observations. Thus each sample from $q(\F_1)q(\F_2)$ corresponds to a composition mapping $\X$ to $\Y$ (different samples correspond to different compositions). This is in contrast to sampling from the factorised distribution of the inducing points (which directly parametrise each $\{f_i\}$). In such case, some compositions (\eg $f_1(x) = -x, f_2(x) = f_3(x) = x$) do not fit the data, making the variational posterior collapse, as argued in Sec.~\ref{sec:mean_field}.
\begin{figure*}[h!]
    \begin{center}
        \hspace{-2cm}
        \includegraphics{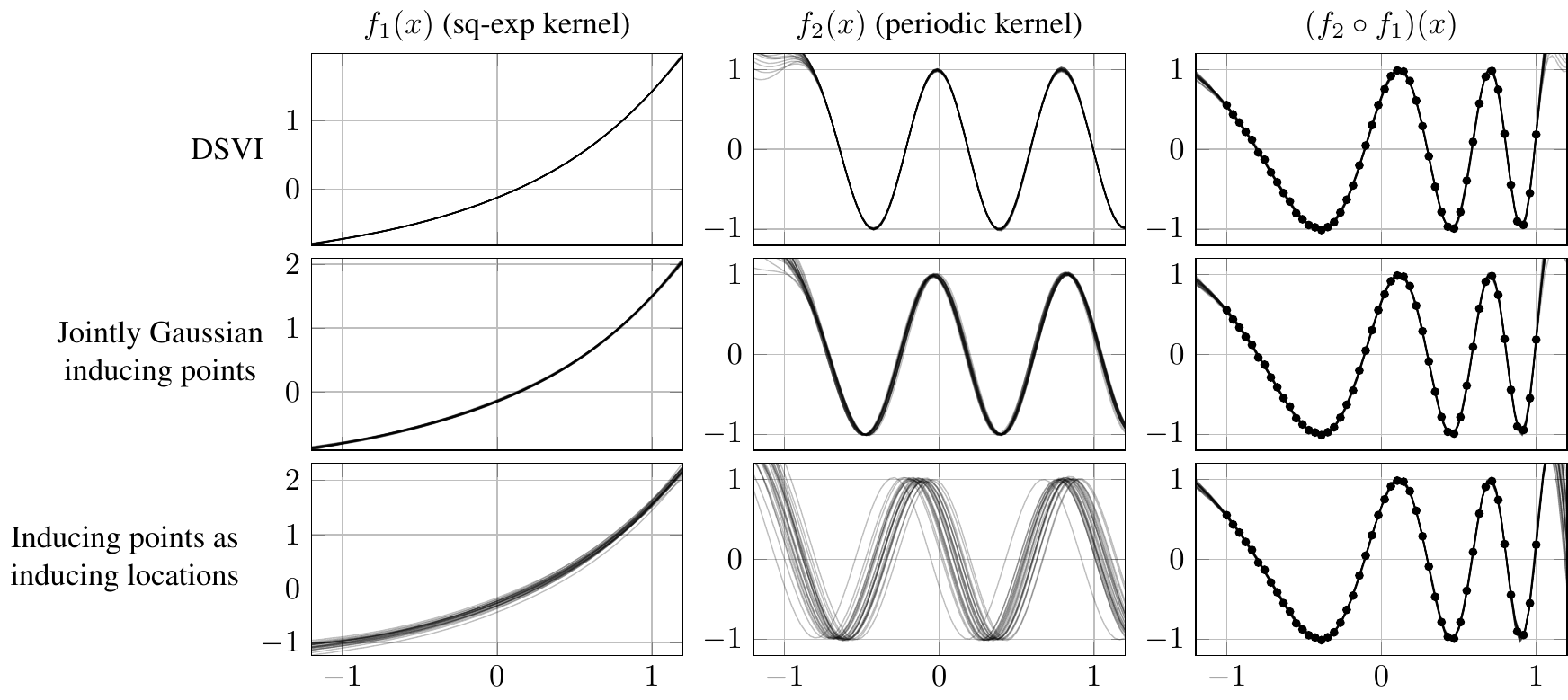}
    \end{center}
    \caption{25 random samples from 2-layer DGPs with squared-exponential and periodic kernels fitted to the observations in the third column (black dots) using DSVI as well as variational distributions discussed in Sec.~\ref{sec:beyond_mean_field}. The first and second columns show samples from each of the two layers, while the third one shows samples from the entire composition (all such samples fit the data despite the variance in $f_1$ and $f_2$ because the two layers are dependent).}
    \label{fig:chirp}
\end{figure*}
\begin{table*}[t]
    \centering
    \begin{tabular}{cccc}
    & ELBO & $\V{f_1(0)}$ & $\V{f_2(0)}$ \\ \midrule
    DSVI & 
        13.43 {\scriptsize $\pm \, 8.03$} & 
        $1.99 \cdot 10^{-6}$ {\scriptsize $\pm \, 1.76 \cdot 10^{-7}$} &
        $1.11 \cdot 10^{-4}$ {\scriptsize $\pm \, 1.35 \cdot 10^{-5}$} \\ 
    Jointly Gaussian &
        23.15 {\scriptsize $\pm \, 6.80$} & 
        $4.23 \cdot 10^{-5}$ {\scriptsize $\pm \, 3.17 \cdot 10^{-6}$} &
        $3.33 \cdot 10^{-4}$ {\scriptsize $\pm \, 2.12 \cdot 10^{-5}$} \\ 
    Inducing points as inducing inputs &
        36.31 {\scriptsize $\pm \, 3.55$} & 
        $2.22 \cdot 10^{-3}$ {\scriptsize $\pm \, 2.73 \cdot 10^{-4}$} &
        $4.98 \cdot 10^{-2}$ {\scriptsize $\pm \, 7.78 \cdot 10^{-3}$} \\ 
    \bottomrule
    \end{tabular}
\caption{Evaluations of the DGPs fitted on a dataset in Fig.~\ref{fig:chirp}. First column shows lower bounds on marginal likelihood $p(\Y)$; the second and third ones show marginal variances of both layers at $x = 0$. The numbers are the means as well as standard deviations across 10 trials.}
\label{table:evals1}
\end{table*}

\paragraph{Inducing inputs} We introduce inducing inputs $\protect{\Z \in \RR^M}$ (with $M < N$ but sufficiently many to satisfy the assumptions outlined in Sec.~\ref{sec:mean_field}) in the input space and denote the evaluations of intermediate layers at $\Z$ as $\Fz_\ell \sim (f_\ell \circ \ldots \circ f_1)(\Z)$. The augmented DGP joint distribution is
\begin{align}
    & p(\Y, \F_L, \ldots, \F_1, \Fz_L, \ldots, \Fz_1 \given \X, \Z) = \\
    & = p(\Y \given \F_L) \prod_{\ell = 1}^L p(\F_\ell \given \Fz_\ell, \F_{\ell-1}, \Fz_{\ell-1}) p(\Fz_\ell \given \Fz_{\ell-1}), \nonumber
\end{align}
where $p(\Fz_\ell \given \Fz_{\ell-1}) \sim \N(\mu_\ell(\Fz_{\ell-1}), k_\ell(\Fz_{\ell-1}, \Fz_{\ell-1}))$ is an $\ell$-th layer GP prior, and $p(\F_\ell \given \Fz_\ell, \F_{\ell-1}, \Fz_{\ell-1})$ is an $\ell$-th layer GP posterior at inputs $\F_{\ell-1}$ given $\Fz_\ell$ and $\Fz_{\ell-1}$ in $\ell$-th and $(\ell-1)$-th layers respectively.

\paragraph{Variational lower bound}

We introduce the following variational distribution
\begin{equation}
    q(\{\F_\ell\}, \{\Fz_\ell\}) = \prod_{\ell = 1}^L p(\F_\ell \given \Fz_\ell, \F_{\ell-1}, \Fz_{\ell-1}) q(\Fz_\ell),
    \label{eq:q_distribution_variational_outputs_inducing_inputs}
\end{equation}
where $q(\Fz_\ell) \sim \N(\m_\ell, \Sb_\ell)$. The corresponding likelihood lower bound is as follows
\begin{align}
    & \mathcal{L}(\Y)  \geq \E_q \left[ \log \frac{p(\Y, \{\F_\ell\}, \{\Fz_\ell\})}{q(\{\F_\ell\}, \{\Fz_\ell\})} \right] = \nonumber \\
    & = \E_{q(\F_L)} [ \log p(\Y \given \F_L)] - \label{eq:variational_outputs_inducing_inputs_lower_bound_1} \\
    & \quad - \sum_{\ell = 1}^L \E_{q(\Fz_\ell)q(\Fz_{\ell-1})} \left[ \log \frac{q(\Fz_\ell)}{p(\Fz_\ell \given \Fz_{\ell-1})} \right]. \label{eq:variational_outputs_inducing_inputs_lower_bound_2}
\end{align}

\paragraph{Estimating \eqref{eq:variational_outputs_inducing_inputs_lower_bound_1}}

We are estimating an expectation over the marginal $q(\F_L) \sim (f_L \circ \ldots \circ f_1)(\X)$, which can be computed by marginalising the intermediate layers in the joint variational posterior \eqref{eq:q_distribution_variational_outputs_inducing_inputs}:
\begin{equation}
    \begin{aligned}
        & q(\F_L) = \int q(\{\F_\ell\}, \{\Fz_\ell\}) \, \dd\{\F_\ell\}_{\ell=1}^{L - 1} \, \dd\{\Fz_i\}_{\ell=1}^{L} \\
        & = \int p(\F_L \given \Fz_L, \F_{L-1}, \Fz_{L-1}) q(\Fz_L) \, \dd\Fz_L \times \\
        & \qquad \times \prod_{\ell=1}^{L - 1} p(\F_\ell \given \Fz_\ell, \F_{\ell-1}, \Fz_{\ell-1}) q(\Fz_\ell) \, \dd\F_\ell \, \dd\Fz_\ell. 
    \end{aligned}
    \label{eq:variational_outputs_predictive}
\end{equation}

The integrals in \eqref{eq:variational_outputs_predictive} are generally intractable since they require integrating the kernel matrices, thus we estimate them by sampling. 
Overall, the procedure is as follows:
\begin{enumerate}[leftmargin=1.5em,itemsep=0pt,topsep=-1pt]
    \item Draw $S$ samples \\ $\{(\F_1^{\Z,s}, \ldots, \F_L^{\Z,s})\}_{s=1}^S \stackrel{iid}{\sim} q(\Fz_1) \cdot \ldots \cdot q(\Fz_L)$,
    \item Use the samples of $\{\Fz_\ell\}$ to sequentially draw samples of intermediate layers $\F^s_\ell \sim p(\F_\ell \given \F_\ell^{\Z, s}, \F^s_{\ell-1}, \F_{\ell-1}^{\Z, s})$ from a GP posterior given $\F_\ell^{\Z, s}$ and $\F_{\ell-1}^{\Z, s}$,
    \item Use $\{\F^s_L\}_{s=1}^s$, the samples from $q(\F_L)$, to estimate the expectation in \eqref{eq:variational_outputs_inducing_inputs_lower_bound_1}: \\
    $\E_{q(\F_L)} [ \log p(\Y \given \F_L)] \approx \frac{1}{S} \sum_{s=1}^S \log p(\Y \given \F^s_L)$.
\end{enumerate}

\paragraph{Estimating \eqref{eq:variational_outputs_inducing_inputs_lower_bound_2}} We write the summands in \eqref{eq:variational_outputs_inducing_inputs_lower_bound_2} as 
\begin{equation}
    \begin{aligned}
        \E&_{q(\Fz_\ell)q(\Fz_{\ell-1})} \left[ \log \frac{q(\Fz_\ell)}{p(\Fz_\ell \given \Fz_{\ell-1})} \right] = \\
        & \quad = \E_{q(\Fz_{\ell-1})} \KL{q(\Fz_\ell)}{p(\Fz_\ell \given \Fz_{\ell-1})}.
    \end{aligned}
    \label{eq:variational_outputs_KL_term}
\end{equation}
KL divergence between the two Gaussians
$q(\Fz_\ell)$ and $p(\Fz_\ell \given \Fz_{\ell-1})$ is a function of $\Fz_{\ell-1}$ and can be computed analytically for a given value of $\Fz_{\ell-1}$. Therefore, to estimate it, we use the draws from $\Fz_{\ell-1}$( which are computed for the estimate of \eqref{eq:variational_outputs_inducing_inputs_lower_bound_1} as well): for every such draw $\F_{\ell-1}^{\Z,s}$, we analytically compute the KL divergence $\KL{q(\Fz_\ell)}{p(\Fz_\ell \given \F_{\ell-1}^{\Z,s})}$, and then average these values to obtain a Monte-Carlo estimate of the expectation in \eqref{eq:variational_outputs_KL_term}.

\paragraph{Learning and predictions} We maximise the likelihood lower bound (\ref{eq:variational_outputs_inducing_inputs_lower_bound_1}-\ref{eq:variational_outputs_inducing_inputs_lower_bound_2}) w.r.t.\ the variational parameters $\{\m_\ell\}$ and $\{\Sb_\ell\}$. The gradients can be obtained using a reparametrisation trick~\citep{Kingma:2013}.
Given a test input $\Xs$, we can draw the DGP outputs $\Fs_L \sim (f_L \circ \ldots \circ f_1)(\Xs)$ by drawing from $q(\Fs_l)$ using the procedure for estimating \eqref{eq:variational_outputs_predictive} described above. We substitute $\Xs$ instead of $\X$ replacing $\F_\ell$ with $\Fs_\ell$ in \eqref{eq:variational_outputs_predictive}, while the rest of the procedure remains the same.

\paragraph{Time complexity} The time complexity of estimating \eqref{eq:variational_outputs_inducing_inputs_lower_bound_1} is $O(LNM^3)$. Sampling from $q(\Fz_i)$ is $O(M^3)$, while, as discussed in \citep{Salimbeni:2017}, sampling from $p(\F_i \given \Fz_i, \F_{i-1}, \Fz_{i-1})$ can be performed separately for each element of $\F_i$ only requiring drawing from univariate Gaussians, which scales linearly with the number of layers and training inputs. The estimate of \eqref{eq:variational_outputs_inducing_inputs_lower_bound_2} does not add additional complexity since we use the samples from $q(\Fz_i)$ drawn for estimating \eqref{eq:variational_outputs_inducing_inputs_lower_bound_1}, while analytic computation of the KL divergence between $q(\Fz_\ell)$ and $p(\Fz_\ell \given \Fz_{\ell-1})$ is $O(M^3)$ since it requires inversions of covariance matrices. Therefore, the overall complexity of estimating the lower bound is $O(LNM^3)$.
\section{NUMERICAL SIMULATIONS} \label{sec:experiments}

\paragraph{Compositional uncertainty} 
As illustrated in Fig.~\ref{fig:chirp} (first row) as well as in Table~\ref{table:evals1}, the intermediate layers in a DGP with a factorised variational distribution over the inducing points collapse to nearly deterministic transformations in the range of the observed data ($[-1, 1]$). Meanwhile, the models with correlated inducing points (second and third rows) capture more uncertainty, with the approach proposed in Sec.~\ref{subsec:variational_intermediate_layers} allowing us to capture more uncertainty than jointly Gaussian inducing points. Additional examples are provided in Appendix~\ref{app:additional-results}.

\paragraph{Likelihood lower bounds}  In Table~\ref{table:evals1} we provide the variational lower bounds of the marginal likelihood\footnote{The baseline estimate of the true marginal likelihood could be obtained by fitting the DGP using HMC~\citep{Havasi:2018}, however, we found the existing implementation of this scheme to be very unstable (as also noted by the authors) and the estimation of marginal likelihood from posterior samples to have high variance, hence we do not report such values.}, $p(\Y)$. We see that including the dependencies between the layers to the variational distribution leads to higher likelihood bounds, suggesting that factorised variational distributions are suboptimal for DGP inference. \vspace{-0.2cm}

\begin{figure*}[t]
    \begin{center}
        \hspace{-2cm}
        \includegraphics{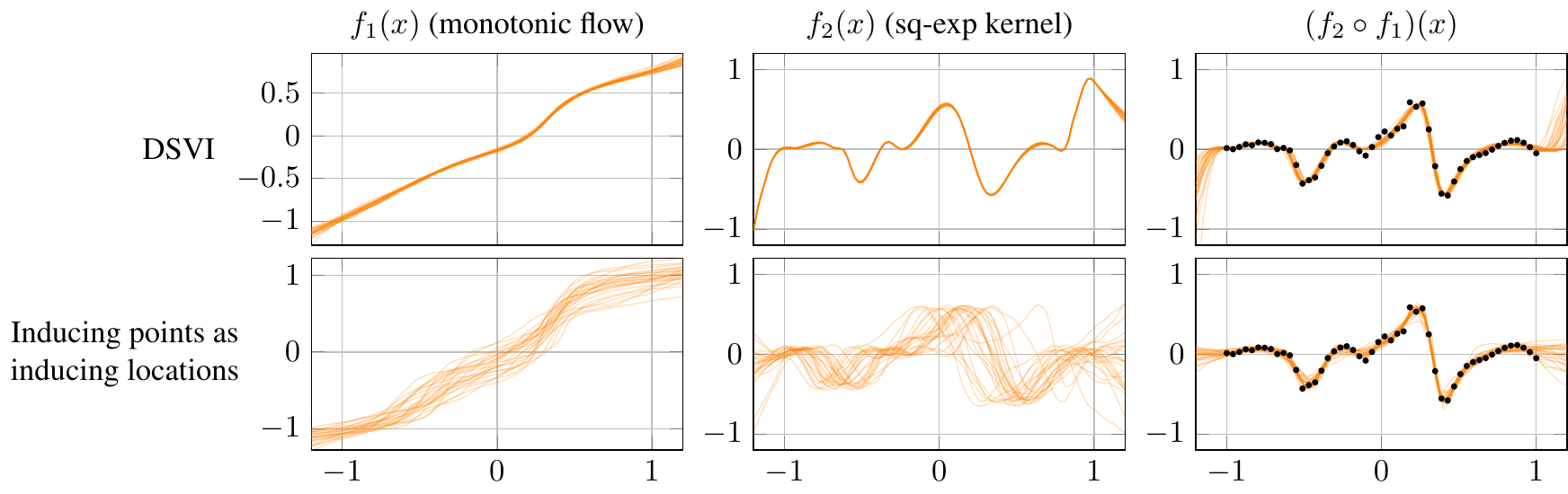}
    \end{center}
    \vspace{-0.cm}
    \caption{Compositional model of heartbeats data, comparing results without (top) and with correlations across layers.}
    \label{fig:heartbeats}
\end{figure*}

\section{APPLICATIONS}
As compositions of functions, DGPs provide a natural way to represent data that is known to have a compositional structure and thus they may be used in applications to learn a more informative representation of the data.

\paragraph{Non-stationary time series}
Consider a sequence $\y \in \RR^N$ that is observed at fixed time inputs $\x \in \RR^N$. 
The observed sequence is assumed to be generated by temporally warping the inputs $\x$ as follows:
\begin{equation}
    \y = f(g(\x)) + \epsilon, \qquad \epsilon \sim \N(0, \sigma^2) \label{eq:nonstationary}
\end{equation} where $g(\cdot)$ is the temporal warping, $f(\cdot)$ is the latent function that encodes the structure of the observed sequence.
The model in \eqref{eq:nonstationary} generates non-stationary sequences, which are convenient to model with a composition of a monotonic transformation of the inputs $\x$ and a GP with a stationary kernel.
The previous work on such models treats the temporal warping $g(\cdot)$ as a deterministic transformation~\citep{Snoek:2014, Kazlauskaite:2019}, disregarding the fact that many different compositions may explain the observed data equally well.

To illustrate this, we consider a recording of a heartbeat~\citep{Bentley:2011}, and fit a two layer DGP with monotonic flow~\citep{Ustyuzhaninov:2020} in the first layer. 
Here the prior on the warping functions $g(\cdot)$ dictates that while an identity warp is preferred, other smooth warps are plausible.
The latent functions $f(\cdot)$ are modelled using a GP with a stationary squared exponential kernel. 
Fig.~\ref{fig:heartbeats} shows how introducing correlations between the layers allows us to uncover a wide range of possible solutions that follow the above-defined priors and are consistent with the data.
Meanwhile, the model with the same prior assumptions that uses a mean-field approximation collapses to a near-deterministic transformation, concentrating the probability mass in both layers on one of the many possible solutions.
An application to sequence alignment is provided in Appendix~\ref{sec:alignment}.

\section{DISCUSSION}

We have discussed the issue of compositional uncertainty in the context of DGPs.
This is in contrast to much of the existing work on DGPs (as well as other Bayesian deep learning approaches~\citep{Gal:2016}) that primarily focuses on predictive uncertainty. 
We argued that the uncertainty about the function implemented by each individual layer in the hierarchy provides a more informative model of the data.
The inference in DGP models is typically performed using variational approximations that factorise across the layers of the hierarchy.  
While computationally convenient, such a factorisation implies that the distributions of
the intermediate layers collapse to deterministic transformations.
Such behaviour diminishes some of the other benefits offered by a compositional model of GPs, such as a systematic way to impose informative functional priors over each of the layers in the hierarchy and a way to uncover distributions over each layer.

To gain further insight into the issue of compositional uncertainty, we proposed two alternatives to the factorised variational distributions of inducing points that include some correlations between the layers.
Contrary to the factorised distributions in DSVI, the proposed variational distributions uncover a range of possible solutions, reinforcing the argument that mean-field approximations are prohibitive when it comes to capturing compositional uncertainty. 
These consideration pose many open questions, ranging from technical considerations of more efficient ways to introduce correlations across layers and ways to represent variational distributions that are multi-modal~\citep{Lawrence:2000}, to broader questions about the structures captured by each layer of the hierarchy, and the applications that may benefit from the more accurate estimates of compositional uncertainty.

\subsubsection*{Acknowledgments}
This work has been supported by EPSRC CDE (EP/L016540/1), CAMERA (EP/M023281/1), EPSRC DTP, Hans Werth\'{e}n Fund at The Royal Swedish Academy of Engineering Sciences, German Federal Ministry of Education and Research (project 01 IS 18049 A) and the Royal Society. 

\bibliography{references}

\begin{thebibliography}{}

\bibitem[Bentley et~al., 2011]{Bentley:2011}
Bentley, P., Nordehn, G., Coimbra, M., and Mannor, S. (2011).
\newblock Pascal {C}lassifying {H}eart {S}ounds {C}hallenge.

\bibitem[Bijl, 2018]{Bijl:2018}
Bijl, H. (2018).
\newblock Lqg and gaussian process techniques: For fixed-structure wind turbine
  control.
\newblock {\em PhD thesis, Delft University of Technology}.

\bibitem[Bui et~al., 2016]{Bui:2016}
Bui, T., Hernandez-Lobato, D., Hernandez-Lobato, J., Li, Y., and Turner, R.
  (2016).
\newblock Deep gaussian processes for regression using approximate expectation
  propagation.
\newblock In {\em International Conference on Machine Learning}.

\bibitem[Cutajar, 2019]{Cutajar:2019}
Cutajar, K. (2019).
\newblock {\em Broadening the scope of Gaussian processes for large-scale
  learning}.
\newblock PhD thesis, Thesis.

\bibitem[Dai et~al., 2016]{Dai:2016}
Dai, Z., Damianou, A., González, J., and Lawrence, N.~D. (2016).
\newblock Variational auto-encoded deep gaussian processes.
\newblock In {\em International Conference on Learning Representations}.

\bibitem[Damianou, 2015]{Damianou:2015}
Damianou, A. (2015).
\newblock Deep gaussian processes and variational propagation of uncertainty.
\newblock {\em PhD Thesis, University of Sheffield}.

\bibitem[Damianou and Lawrence, 2013]{Damianou:2013}
Damianou, A. and Lawrence, N. (2013).
\newblock Deep gaussian processes.
\newblock In {\em International Conference on Artificial Intelligence and
  Statistics (AISTATS)}.

\bibitem[{Der Kiureghian} and Ditlevsen, 2009]{DerKiureghian:2009}
{Der Kiureghian}, A. and Ditlevsen, O. (2009).
\newblock Aleatoric or epistemic? does it matter?
\newblock {\em Structural Safety}, 31.

\bibitem[Dunlop et~al., 2018]{Dunlop:2018}
Dunlop, M.~M., Girolami, M.~A., Stuart, A.~M., and Teckentrup, A.~L. (2018).
\newblock How deep are deep gaussian processes?
\newblock {\em Journal of Machine Learning Research}.

\bibitem[Duvenaud et~al., 2014]{Duvenaud:2014}
Duvenaud, D., Rippel, O., Adams, R.~P., and Ghahramani, Z. (2014).
\newblock Avoiding pathologies in very deep networks.
\newblock In {\em International Conference on Artificial Intelligence and
  Statistics (AISTATS)}.

\bibitem[Gal, 2016]{Gal:2016}
Gal, Y. (2016).
\newblock {\em Uncertainty in Deep Learning}.
\newblock PhD thesis, University of Cambridge.

\bibitem[Gal and Ghahramani, 2016]{Gal:2016:2}
Gal, Y. and Ghahramani, Z. (2016).
\newblock Dropout as a bayesian approximation: Representing model uncertainty
  in deep learning.
\newblock In {\em International Conference on Machine Learning}.

\bibitem[Girard et~al., 2003]{Girard:2003}
Girard, A., Rasmussen, C.~E., Candela, J.~Q., and Murray-Smith, R. (2003).
\newblock Gaussian process priors with uncertain inputs application to
  multiple-step ahead time series forecasting.
\newblock In {\em Neural Information Processing Systems}.

\bibitem[Havasi et~al., 2018]{Havasi:2018}
Havasi, M., Hern\'{a}ndez-Lobato, J.~M., and Murillo-Fuentes, J.~J. (2018).
\newblock Inference in deep gaussian processes using stochastic gradient
  hamiltonian monte carlo.
\newblock In {\em Neural Information Processing Systems}.

\bibitem[Hensman et~al., 2017]{Hensman:2017}
Hensman, J., Durrande, N., and Solin, A. (2017).
\newblock Variational fourier features for gaussian processes.
\newblock {\em Journal of Machine Learning Research (JMLR)}, 18(1).

\bibitem[Hensman et~al., 2013]{Hensman:2013}
Hensman, J., Fusi, N., and Lawrence, N.~D. (2013).
\newblock {G}aussian processes for big data.
\newblock In {\em Conference on Uncertainty in Artificial Intelligence (UAI)}.

\bibitem[Hensman and Lawrence, 2014]{Hensman:2014}
Hensman, J. and Lawrence, N.~D. (2014).
\newblock Nested variational compression in deep {G}aussian processes.
\newblock {\em arXiv preprint arXiv:1412.1370}.

\bibitem[Jin et~al., 2017]{Jin:2017}
Jin, M., Damianou, A., Abbeel, P., and Spanos, C. (2017).
\newblock Inverse reinforcement learning via deep gaussian process.
\newblock {\em Conference on Uncertainty in Artificial Intelligence (UAI)}.

\bibitem[Kaiser et~al., 2018]{Kaiser:2018}
Kaiser, M., Otte, C., Runkler, T., and Ek, C.~H. (2018).
\newblock Bayesian alignments of warped multi-output gaussian processes.
\newblock In {\em Neural Information Processing Systems}.

\bibitem[Kazlauskaite et~al., 2019]{Kazlauskaite:2019}
Kazlauskaite, I., Ek, C.~H., and Campbell, N. (2019).
\newblock Gaussian process latent variable alignment learning.
\newblock In {\em International Conference on Artificial Intelligence and
  Statistics (AISTATS)}.

\bibitem[Kingma and Welling, 2014]{Kingma:2013}
Kingma, D.~P. and Welling, M. (2014).
\newblock Auto-encoding variational bayes.
\newblock In {\em International Conference on Learning Representations}.

\bibitem[Lawrence, 2000]{Lawrence:2000}
Lawrence, N.~D. (2000).
\newblock {\em Variational Inference in Probabilistic Models}.
\newblock PhD thesis, Cambridge University.

\bibitem[Lawrence, 2004]{Lawrence:2004}
Lawrence, N.~D. (2004).
\newblock Gaussian process latent variable models for visualisation of high
  dimensional data.
\newblock {\em Neural Information Processing Systems}.

\bibitem[Lawrence and Moore, 2007]{Lawrence:2007}
Lawrence, N.~D. and Moore, A.~J. (2007).
\newblock Hierarchical gaussian process latent variable models.
\newblock In {\em International Conference on Machine Learning}.

\bibitem[L\'{a}zaro-Gredilla, 2012]{Lazaro:2012}
L\'{a}zaro-Gredilla, M. (2012).
\newblock Bayesian warped gaussian processes.
\newblock In {\em Neural Information Processing Systems}.

\bibitem[Mchutchon and Rasmussen, 2011]{Mchutchon:2011}
Mchutchon, A. and Rasmussen, C.~E. (2011).
\newblock Gaussian process training with input noise.
\newblock In {\em Neural Information Processing Systems}.

\bibitem[Rainforth et~al., 2019]{Rainforth:2019}
Rainforth, T., Cornish, R., Yang, H., Warrington, A., and Wood, F. (2019).
\newblock On nesting monte carlo estimators.
\newblock {\em Proceedings of Machine Learning Research}, 80.

\bibitem[Rasmussen and Williams, 2005]{Rasmussen:2005}
Rasmussen, C.~E. and Williams, C. K.~I. (2005).
\newblock {\em {Gaussian Processes for Machine Learning}}.
\newblock MIT Press.

\bibitem[Salimbeni and Deisenroth, 2017]{Salimbeni:2017}
Salimbeni, H. and Deisenroth, M. (2017).
\newblock Doubly stochastic variational inference for deep gaussian processes.
\newblock In {\em Neural Information Processing Systems}.

\bibitem[Snoek et~al., 2014]{Snoek:2014}
Snoek, J., Swersky, K., Zemel, R., and Adams, R. (2014).
\newblock Input warping for bayesian optimization of non-stationary functions.
\newblock In {\em International Conference on Machine Learning}.

\bibitem[Sun et~al., 2019]{Sun:2019}
Sun, S., Zhang, G., Shi, J., and Grosse, R. (2019).
\newblock Functional variational bayesian neural networks.
\newblock In {\em International Conference on Learning Representations}.

\bibitem[Titsias and Lawrence, 2010]{Titsias:2010}
Titsias, M. and Lawrence, N. (2010).
\newblock Bayesian gaussian process latent variable model.
\newblock {\em Journal of Machine Learning Research (JMLR)}, 9.

\bibitem[Ustyuzhaninov et~al., 2020]{Ustyuzhaninov:2020}
Ustyuzhaninov, I., Kazlauskaite, I., Ek, C.~H., and Campbell, N. D.~F. (2020).
\newblock Monotonic gaussian process flow.
\newblock In {\em International Conference on Artificial Intelligence and
  Statistics (AISTATS)}.

\end{thebibliography}
\bibliographystyle{apalike}

\clearpage
\appendix
\renewcommand\thefigure{\thesection\arabic{figure}}

\section{Derivation of a counterexample in Sec.~\ref{sec:mean_field}}
\label{app:counterexample}
\setcounter{figure}{0}

Our derivations follow~\citep{Girard:2003}, who study GPs with uncertain inputs. 
Specifically they compute the mean and the variance of $f(x_*)$, where $f \sim \GP(\mu(\cdot), k(\cdot, \cdot))$, and $x_* \sim \N(\mu_*, \sigma_*^2)$.

According to Eq.~(12) in~\citet{Girard:2003},
\begin{equation}
    \begin{aligned}
        v(\mu_*, \sigma_*^2) 
           :&= \V{f(x_*)} \\
            & = 1 + \Tr\left[(\beta \beta^T - \K^{-1})Q \right] - \Tr(\q^T \beta)^2,
    \end{aligned}
    \label{eq:app:variance-uncertain-inputs}
\end{equation}
where $\K_{ij} = k(\Z_i, \Z_j)$ is a kernel matrix at inducing locations. Assuming for simplicity a squared-exponential kernel $k(x, x') = \exp((x - x')^2 / 2\gamma^2)$, and a single inducing point $(z, u)$, $\K$ becomes a scalar, $\K = 1$. The matrix $Q$ in the equation above has as many rows and columns as there are inducing points, meaning that under our assumptions, $Q$ is a scalar given by the following equation:
\begin{equation*}
    \begin{aligned}
        Q = \frac{1}{\sqrt{\frac{2\sigma_*^2}{\gamma^2} + 1}} 
            \exp \left( 
                - \frac{1}{2 \left(\frac{\gamma^2}{2} + \sigma_*^2 \right)} (u - \mu_*)^2
            \right).
    \end{aligned}
\end{equation*}

The term $\beta$ in \eqref{eq:app:variance-uncertain-inputs} is defined as $\beta = \K^{-1} \U$, which equals zero assuming that the single inducing point $u$ is equal to zero.

In summary, under our assumptions,
\begin{equation}
    v(\mu_*, \sigma_*^2) := \V{f(x_*)} = 1 - Q.
    \label{eq:app:variance-uncertain-inputs-reduced}
\end{equation}

The derivative of \eqref{eq:app:variance-uncertain-inputs-reduced} w.r.t. $\sigma_*^2$ is as follows
\begin{equation*}
    \begin{aligned}
        & \frac{\partial v(\mu_*, \sigma_*^2)}{\partial \sigma_*^2}
            = \exp \left( 
                - \frac{1}{2 \left(\frac{\gamma^2}{2} + \sigma_*^2 \right)} (u - \mu_*)^2
            \right) \times \\
            & \quad \times \left(-
                \frac{(u - \mu_*)^2}{2\left(\frac{\gamma^2}{2} + \sigma_*^2 \right)^2}
                +\frac{1}{\gamma^2 \left( \frac{2\sigma_*^2}{\gamma^2} + 1 \right)^{3/2}} 
            \right).
    \end{aligned}
\end{equation*}

Evaluating this derivative at $\sigma_*^2 = 0$, we obtain
\begin{equation*}
    \begin{aligned}
        & \left. \frac{\partial v(\mu_*, \sigma_*^2)}{\partial \sigma_*^2} \right\rvert_{\sigma_*^2 = 0}
            = \exp \left( 
                - \frac{(u - \mu_*)^2}{\gamma^2}
            \right) \times \\
            & \quad \times \left(
                -\frac{2 (u - \mu_*)^2}{\gamma^4}
                +\frac{1}{\gamma^2} 
            \right).
    \end{aligned}
\end{equation*}

From the above equation it is easy to see that
\begin{equation*}
    \left. \frac{\partial v(\mu_*, \sigma_*^2)}{\partial \sigma_*^2} \right\rvert_{\sigma_*^2 = 0} < 0 \quad \Leftrightarrow \quad \gamma < \sqrt{2} |u - \mu_*|.
\end{equation*}

In other words, if the input mean is sufficiently far away from the inducing point (in relation to the length scale), \ie $\gamma < \sqrt{2} |u - \mu_*|$, adding input noise may reduce the output uncertainty. 

\section{Analytic marginalisation of jointly Gaussian inducing points}
\label{app:analytic-marginalisation}

In this section, we provide the derivation of the variational distribution with analytically marginalised inducing points that have a joint Gaussian distribution, as described in Sec.~\ref{subsec:joint_gaussian_inducing_points}. We first derive the result for a 2-layer case and then discuss a way to generalise beyond two layers.

We consider jointly Gaussian inducing points as
\begin{equation}
    q(\U_1, \U_2) \sim \N\left(
        \begin{pmatrix}
            \m_1 \\ \m_2
        \end{pmatrix},
        \begin{pmatrix}
            S_{11} & S_{12} \\
            S_{21} & S_{22}
        \end{pmatrix}
    \right),
\end{equation}
and a joint variational distribution of a two-layer DGP (suppressing the dependence on the inducing locations $\Z_1$ in the notation) is
\begin{equation}
\begin{aligned}
    q(\F_2, \U_2, \F_1, \U_1) = &p(\F_2 \given \U_2, \F_1) q(\U_2 \given \U_1) \\
    &p(\F_1 \given \U_1, \X) q(\U_1).
    \label{eq:joint-variational-2-layers}
\end{aligned}
\end{equation}

The goal is to integrate $\U_1$ and $\U_2$ out from \eqref{eq:joint-variational-2-layers} in order to fit the model without sampling the inducing points. The following derivations are based on the argument that the mean of $F_i$ is a linear transformation of $U_i$, and vice versa.

Assume that $q(\U_1) \sim \N(\m_1, S_{11})$ and $p(\F_1 \given \X) \sim \N(\tilde{\mu}_1, \tilde{\Sigma}_1)$ with 
\begin{equation}
    \begin{aligned}
        \tilde{\mu}_1 & = \mu_1(\X) + \alpha_1(\X)^T (\m_1 - \mu_1(\X)), \\
        \tilde{\Sigma}_1 & = K_1(\X, \X) - \alpha_1(\X)^T (K_1(\Z_1, \Z_1) - S_{11}) \, \alpha_1(\X), 
    \end{aligned}
\end{equation} where $\alpha_1(\X) = K_1(\Z_1, \Z_1)^{-1} K_1(\Z_1, \X)$.
We can compute the joint distribution $q(\U_1, \F_1 \given \X) = q(\U_1) p(\F_1 \given \U_1, \X)$ using a standard result\footnote{See, for example, Section 4 in \url{https://davidrosenberg.github.io/mlcourse/in-prep/multivariate-gaussian.pdf}.} for a linear model with a Gaussian prior and likelihood (in the following we will be referring to this result as $(*)$) as follows:
\begin{equation}
        q(\U_1, \F_1) \sim \N \left(
            \begin{pmatrix}
                \m_1 \\
                \tilde{\mu}_1
            \end{pmatrix},
            \begin{pmatrix}
                S_{11} & S_{11} \alpha_1(\X) \\
                \alpha_1(\X)^T S_{11} & \tilde{\Sigma}_1
            \end{pmatrix}
        \right).
\end{equation}
From this we can swap $\U_1$ and $\F_1$ in the conditional distribution by computing
\begin{equation}
    \begin{aligned}
        q(\U_1 \given \F_1) \sim \N(& \m_1 + S_{11} \alpha_1(\X) \tilde{\Sigma}_1^{-1} (\F_1 - \tilde{\mu}_1), \\
        & S_{11} - S_{11}\alpha_1(\X) \tilde{\Sigma}_1^{-1} \alpha_1(\X)^T S_{11}).
    \end{aligned}
\end{equation}

Now we can integrate $\U_1$ from \eqref{eq:joint-variational-2-layers} by applying (*) again, obtaining
\begin{align}
    q(\F_2, \U_2, \F_1) & = p(\F_2 \given \U_2, \F_1) q(\U_2 \given \F_1) p(\F_1 \given \X), \\
    \text{where} \label{eq:joint-variational-2-layers-no-u1} \\
    q(\U_2 \given \F_1) & = \N(\m_2 + S_{21} \alpha_1(\X) \tilde{\Sigma}_1^{-1} (\F_1 - \tilde{\mu}_1), \nonumber \\
    & \qquad S_{22} - S_{21} \alpha_1(\X) \tilde{\Sigma}_1^{-1} \alpha_1(\X)^T S_{12}). \nonumber
\end{align}

Another application of (*) allows us to integrate $\U_2$ from \eqref{eq:joint-variational-2-layers-no-u1} obtaining joint distribution of intermediate layers $q(\F_2, \F_1 \given \X) = q(\F_2 \given \F_1) q(\F_1 \given \X)$ with $q(\F_2 \given \F_1) = \N(\tilde{\mu}_2, \tilde{\Sigma}_2)$ where
\begin{equation}
    \begin{aligned}
        \tilde{\mu}_2 & = \mu_2(\F_2) + \alpha_2(\F_1)^{T} (\m_2 + S_{21} \alpha_1(\X) \tilde{\Sigma}_1^{-1} \\
        & \qquad (\F_1 - \tilde{\mu}_1 - \alpha_1(\X)^T (\m_1 - \mu_1(\Z_1)) - \mu_2(\Z_2)), \\
        \tilde{\Sigma}_2 & = K_2(\F_1, \F_1) - \alpha_2(\F_1)^T (K_2(\Z_2, \Z_2) - S_{22} + \\
        & \qquad + S_{21} \alpha_1(\X) \tilde{\Sigma}_1^{-1} \alpha_1(\X)^T S_{12}) \alpha_2(\F_1).
    \end{aligned} \label{eq:joint-dist-inducing-integrate}
\end{equation}

This result can be generalised to more than two layers, starting with  $q(F_i | F_{i-1}, ..., F_1) \sim N(\mu_i, \Sigma_i)$, and repeating the steps outlined above to arrive at $q(F_{i+1} | F_i, F_{i-1}, ..., F_1)$, parameters of which can be deduced by replacing the indices for the first layer with indices for the i$^{\text{th}}$ layer in \eqref{eq:joint-dist-inducing-integrate}. This leads to the result given in \eqref{eq:integrated-mean} and~\eqref{eq:integrated-var}. 

\section{Implementation} 
\setcounter{figure}{0} 

Our implementations for the approaches discussed in Sec.~\ref{subsec:joint_gaussian_inducing_points} and Sec.~\ref{subsec:variational_intermediate_layers} are built on the Tensorflow~(Abadi et al., 2015) and the Tensorflow Probability~(Dillon et al., 2017) libraries.

\section{Alignment task}
\label{sec:alignment}
\setcounter{figure}{0} 

Another example of a task that calls for an explicit representation of the constituent functions is the task of aligning temporal sequences~\citep{Kaiser:2018, Kazlauskaite:2019}.
Consider a set of sequences $\{\y_j\}_{j=1}^J$ where each sequence $\y_j \in \RR^N$ is observed at fixed inputs $\x \in \RR^N$ that typically correspond to time. 
It is known that the observed sequences were generated by temporally warping the inputs $\x$ as follows:
\begin{equation}
    \y_j = f_j(g_j(\x)) + \epsilon_j \label{eq:alignment}
\end{equation} where $g_j(\cdot)$ is the temporal warping, $f_j(\cdot)$ is the latent function that encodes the structure of the observed sequence (that is not corrupted by the temporal warping) and $\epsilon_j \sim \N(0, \sigma^2_j)$ is the observation noise.
Given this construction, prior knowledge may be imposed on the two functions that make up the model; for example, the temporal warps are typically constrained to be monotonic increasing to ensure that the order of observations is preserved, while the latent functions may be described using a Gaussian process prior with an appropriate kernel that represent our beliefs about the features of these functions.
The goal in an alignment task is to learn the model of the data as defined in~\eqref{eq:alignment} such that the latent functions $\{f_j\}$ for all $J$ sequences are as similar as possible, \ie we are interested in such a composition of the functions $f_j$ and $g_j$ such that $\sum_{i=1}^J \sum_{k=i+1}^J (f_i(\x) - f_k(\x))$ (the pairwise distance between the latent functions) is as small as possible given the prior assumptions on $\{f_j\}$ and $\{g_j\}$.
The composition in~\eqref{eq:alignment} can be expressed using a two-layer DGP with appropriate priors (for a detailed description of imposing monotonicity constraints, see~\citep{Ustyuzhaninov:2020}).

\begin{figure*}
	\includegraphics[width=1.\textwidth]{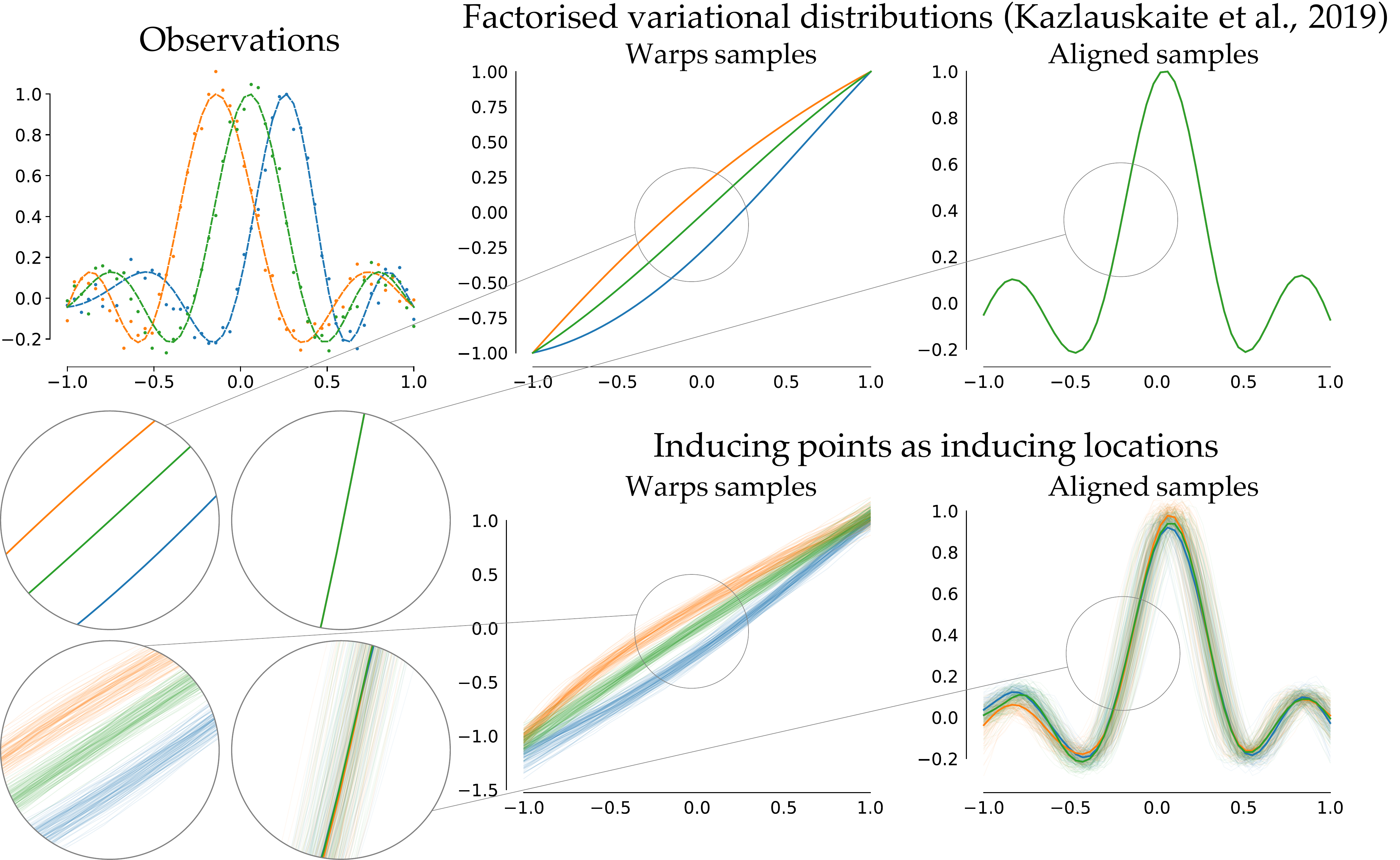}
	\caption{Alignment task. The top left figure shows the observed data that needs to be aligned. The two rows on the right show the alignment used in~\cite{Kazlauskaite:2019}, that provides a point estimate of the solution (top row), and the alignment using a probabilistic model with correlated warping functions and latent functions (bottom row).} \label{fig:align}
\end{figure*}

Consider a set of 3 sequences generated using a sinc function in the range $[-1, 1]$ that need to be aligned. 
Fig.~\ref{fig:align} illustrates how correlations between layers allow us to uncover a set of solutions, as opposed to a point estimate of the warping and the latent functions reported in~\citep{Kazlauskaite:2019}.

Some additional correlations need to be introduced into the alignment model to ensure that any given sample of the 3 latent functions $f_j(\x), j = 1, 2, 3$ at fixed inputs $\x$ are consistent (otherwise, the solution collapses to a single latent function for all sequences which is at odds with our goal of finding a range of possible solutions).
In this example, the additional correlations are introduced by jointly sampling the inducing points that define the first layer of the composition.

\section{Additional numerical simulations}
\label{app:additional-results}
\setcounter{figure}{0} 

\begin{figure*}[t]
	\includegraphics[width=1.\textwidth]{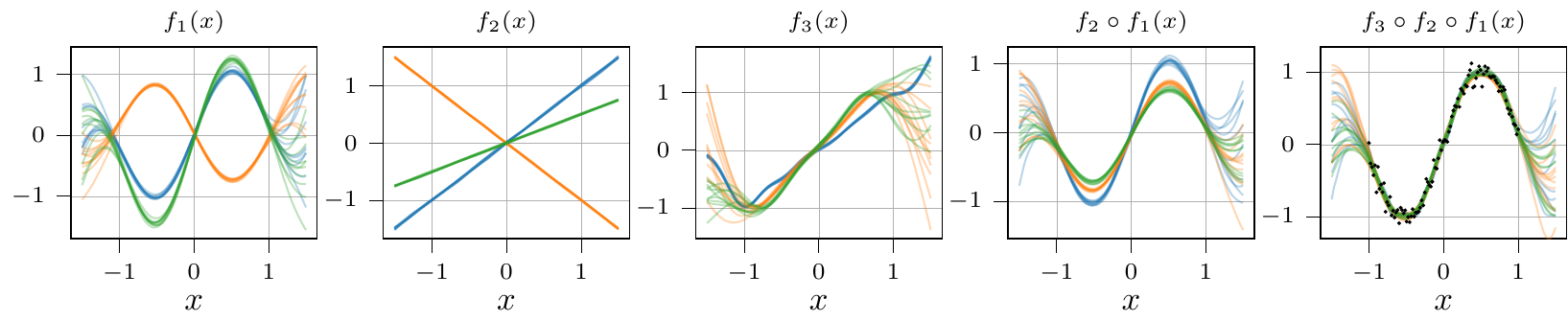}
	\caption{Example fits of a three-layer DGP \emph{with factorised inducing points} to a data set shown in the rightmost panel (black dots). Different panels show the computations performed by each of the three layers and their compositions. Different colours correspond to three models fitted to the same data with different random initialisations. For each initialisation, ten samples (of the same colour) from the fitted model are shown on top of each other.}
	\label{fig:sine_factorised}
\end{figure*}

\begin{figure*}[t]
	\includegraphics[width=1.\textwidth]{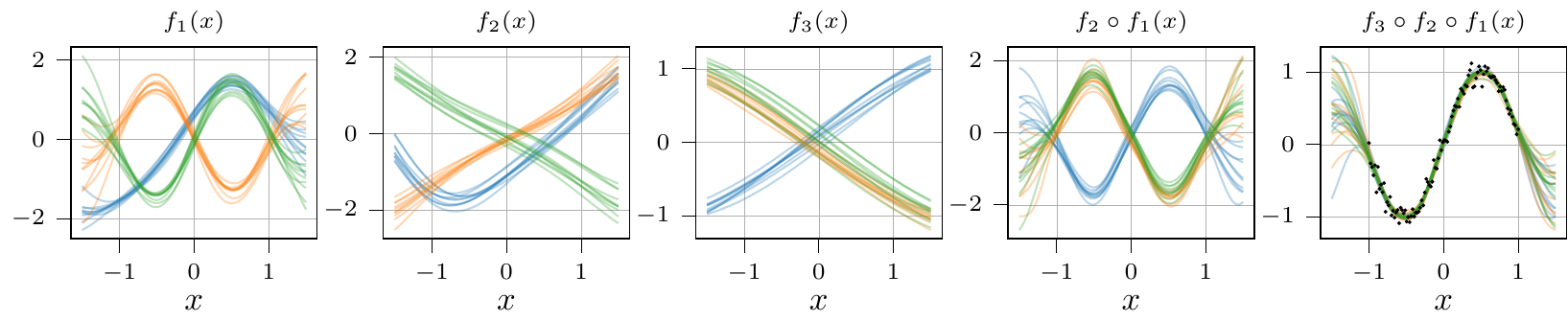}
	\caption{Example fits of a three-layer DGP with \emph{jointly Gaussian inducing points} (Sec.~\ref{subsec:joint_gaussian_inducing_points}). The figure arrangement is the same as in Fig.~\ref{fig:sine_factorised}.}
    \label{fig:sine_gaussian}
\end{figure*}

\begin{figure*}[t]
	\includegraphics[width=1.\textwidth]{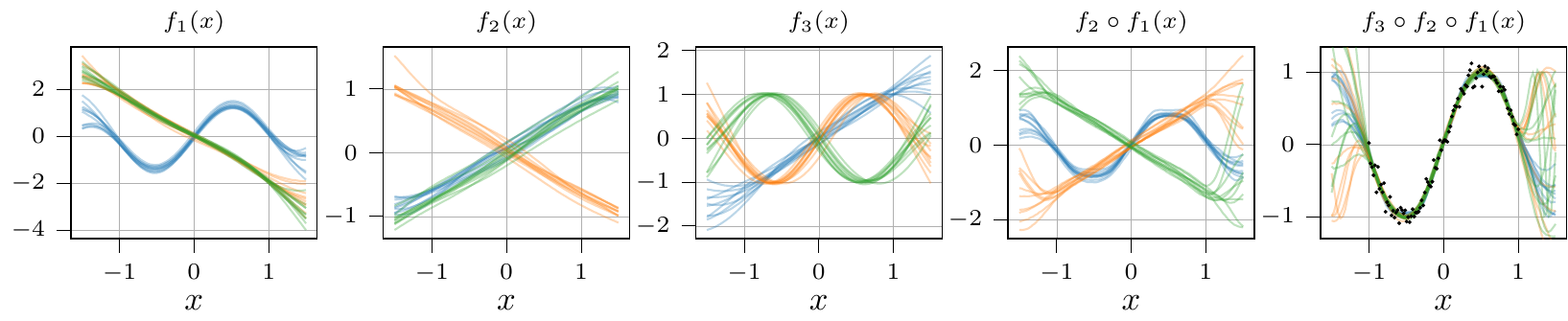}
	\caption{Example fits of a three-layer DGP with \emph{inducing points as inducing locations} (Sec.~\ref{subsec:variational_intermediate_layers}). The figure arrangement is the same as in Fig.~\ref{fig:sine_factorised}}
    \label{fig:sine_variational}
\end{figure*}

\begin{figure*}[t]
	\includegraphics[width=1.\textwidth]{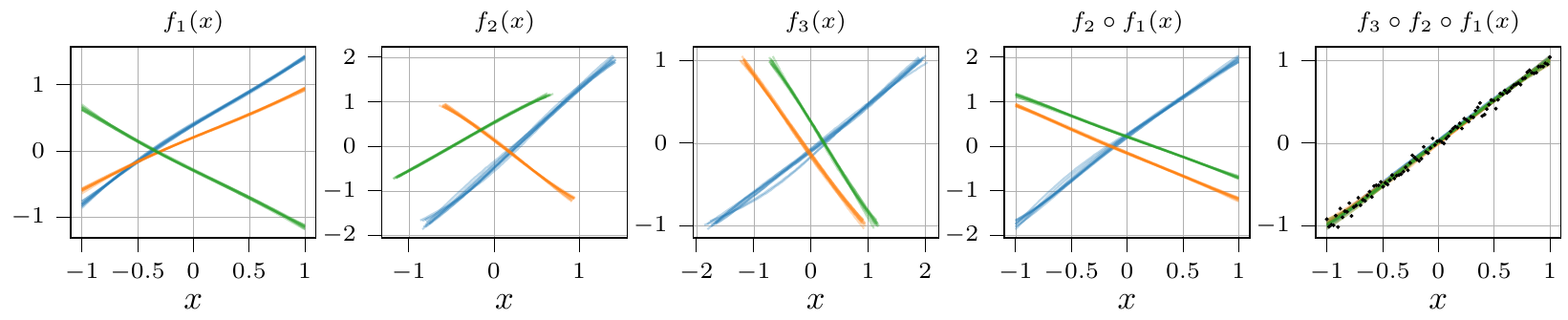}
	\caption{Example fits of a three-layer DGP \emph{with factorised inducing points} to a data set shown in the rightmost panel (black dots). Different panels show the computations performed by each of the three layers and their compositions. Different colours correspond to three models fitted to the same data with different random initialisations. For each initialisation, ten samples (of the same colour) from the fitted model are shown on top of each other.}
	\label{fig:identity_factorised}
\end{figure*}

\begin{figure*}[t]
	\includegraphics[width=1.\textwidth]{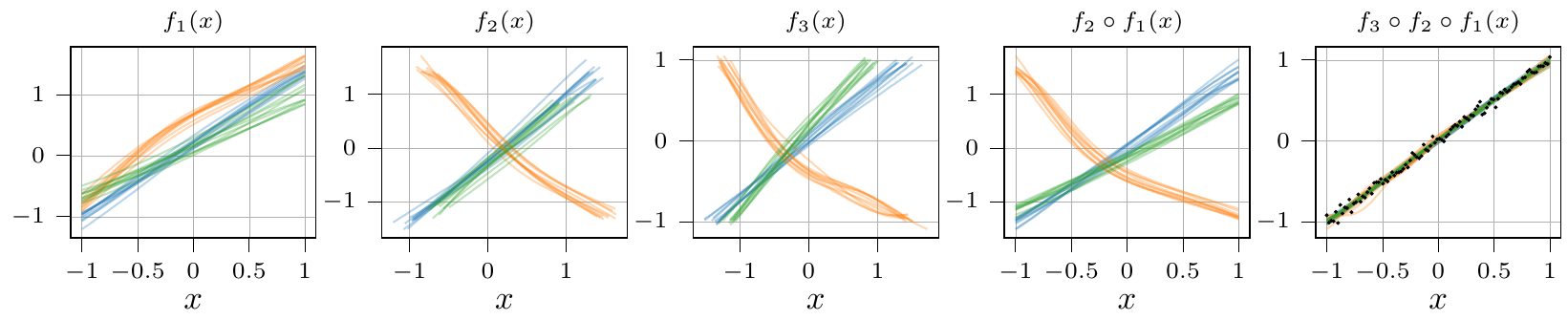}
	\caption{Example fits of a three-layer DGP with \emph{jointly Gaussian inducing points} (Sec.~\ref{subsec:joint_gaussian_inducing_points}). The figure arrangement is the same as in Fig.~\ref{fig:identity_factorised}.}
    \label{fig:identity_gaussian}
\end{figure*}

\begin{figure*}[t]
	\includegraphics[width=1.\textwidth]{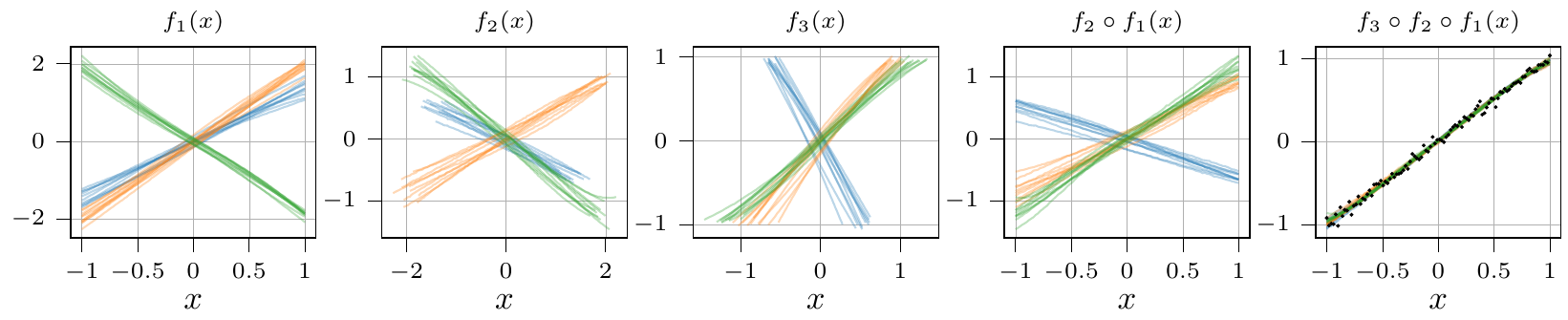}
	\caption{Example fits of a three-layer DGP with \emph{inducing points as inducing locations} (Sec.~\ref{subsec:variational_intermediate_layers}). The figure arrangement is the same as in Fig.~\ref{fig:identity_factorised}}
    \label{fig:identity_variational}
\end{figure*}

In this section we provide additional examples (Fig.~\ref{fig:sine_factorised} to \ref{fig:identity_variational}) of 3-layer DGP fits to two functions, a sine and an identity function. Similar to Fig.~\ref{fig:chirp}, we fit a DGP to both functions using three variational inference schemes based on a factorised variational distribution of inducing points (DSVI), jointly Gaussian inducing points of Sec.~\ref{subsec:joint_gaussian_inducing_points}, and the distribution discussed in Sec.~\ref{subsec:variational_intermediate_layers}.


\clearpage


\clearpage

\section*{References (for appendix)}
\hangindent=0.4cm
M.~Abadi, 
    A.~Agarwal, 
    P.~Barham, 
    E.~Brevdo, 
    Zh.~Chen, 
    C.~Citro, 
    G.~S.~Corrado, 
    A.~Davis, 
    J.~Dean, 
    M.~Devin, 
    S.~Ghemawat, 
    I.~Goodfellow, 
    A.~Harp, 
    G.~Irving, 
    M.~Isard, 
    Y.~Jia, 
    R.~Jozefowicz, 
    L.~Kaiser, 
    M.~Kudlur, 
    J.~Levenberg, 
    D.~Man\'{e}, 
    R.~Monga, 
    Sh.~Moore, 
    D.~Murray, 
    Ch.~Olah, 
    M.~Schuster, 
    J.~Shlens, 
    B.~Steiner, 
    I.~Sutskever, 
    K.~Talwar, 
    P.~Tucker, 
    V.~Vanhoucke, 
    V.~Vasudevan, 
    F.~Vi\'{e}gas, 
    O.~Vinyals, 
    P.~Warden, 
    M.~Wattenberg, 
    M.~Wicke, 
    Y.~Yu, 
    X.~Zheng. (2015)
\textit{TensorFlow: Large-scale machine learning on heterogeneous systems.} Software available from \textit{tensorflow.org.}

\hangindent=0.4cm
J. V.~Dillon,  I.~Langmore,  D. Tran,  E. Brevdo,  S. Vasudevan,  D. Moore,  B. Patton,  A. Alemi,  M. Hoffman,  R. A. Saurous (2017). \textit{TensorFlow Distributions. arXiv preprint arXiv:1711.10604}

\end{document}